%% file: preprint.tex
\documentclass[11pt,letterpaper]{article}
\usepackage[margin=1in]{geometry}
\usepackage[T1]{fontenc}
\usepackage{newtxtext,newtxmath}
\usepackage[hyphens]{url}
\usepackage{graphicx}
\usepackage[round,authoryear]{natbib}
\usepackage{caption}
\usepackage{subcaption}
\usepackage{float}
\usepackage{amsmath}
\usepackage[table]{xcolor}
\usepackage{colortbl}
\usepackage{microtype}
\usepackage{fancyhdr}
\usepackage[hidelinks]{hyperref}
\hypersetup{
  pdftitle={Mitigating Class-Tail Undercoverage in Medical Vision-Language Models under Clinical Shift},
  pdfauthor={Mushir Akhtar and M. Tanveer},
  pdfsubject={Class-tail conformal prediction and selective deferral for medical vision-language models},
  pdfkeywords={medical vision-language models, conformal prediction, class-tail coverage, clinical shift, selective deferral}
}
\newtheorem{proposition}{Proposition}
\definecolor{CalCoDE-Row}{HTML}{D8DEDA}
\newcommand{\best}[1]{\textbf{#1}}
\newcommand{\second}[1]{#1}
\newcommand{\backbonerow}[1]{\hline\multicolumn{7}{l}{\textbf{#1}}\\\hline}
\fancypagestyle{preprintfirst}{%
  \fancyhf{}
  \fancyfoot[L]{\footnotesize\itshape Preprint}
  \fancyfoot[C]{\thepage}

}

\title{\textbf{Mitigating Class-Tail Undercoverage in Medical Vision-Language Models under Clinical Shift}}
\author{%
Mushir Akhtar \qquad M.\ Tanveer\\[0.45em]
\normalsize Department of Mathematics, Indian Institute of Technology Indore, India\\
\normalsize \texttt{mushirakhtar.ml@gmail.com} \qquad
\texttt{mtanveer@iiti.ac.in}}
\date{}

\begin{document}

\maketitle
\thispagestyle{preprintfirst}

\begin{abstract}
Medical vision-language models (VLMs) can retain high observed marginal coverage after clinical shift while substantially under-covering an individual disease class. The affected class varies with acquisition protocol and backbone geometry, so source prevalence does not reliably reveal the failure. Existing localized and tail-aware conformal methods respectively adapt to test neighborhoods and source-frequency tails, leaving held-out class-wise coverage failure unmodeled. We introduce Class-Tail Adaptive Localized Conformal Deferral (CALCoDe), a post-hoc reliability layer for frozen medical VLMs. Cross-fitted validation predictions identify classes at risk of undercoverage, and a disjoint calibration split estimates their class-conditional tail thresholds. CALCoDe combines each protected threshold with a localized conformal threshold using a one-sided maximum. The resulting set contains every label admitted by the localized rule, with additional protection confined to validation-identified classes. An independently calibrated support audit defers cases with insufficient inlier support. Under exchangeability among accepted examples within each protected class, CALCoDe provides finite-sample coverage at the prespecified guard level and contains the corresponding localized conformal sets; coverage on shifted external cohorts is evaluated empirically. Among standard conformal baselines and recent VLM-specific conformal methods evaluated across two dermatology shifts (HAM10000$\rightarrow$ISIC 2019 and HAM10000$\rightarrow$PAD-UFES-20) and four frozen VLM backbones (BiomedCLIP, OpenAI CLIP ViT-B/32, PubMedCLIP ViT-B/32, and MedSigLIP-448), CALCoDe is the only approach whose observed marginal and worst-class accepted coverage both reach 0.95 in all eight settings. On HAM10000$\rightarrow$ISIC 2019, its average worst-class accepted coverage is 0.970, compared with 0.926 for sTACP and 0.864 for LCP-VLM.
\end{abstract}

\section{Introduction and Motivation}

Medical image classification increasingly relies on foundation models that align images with clinical text. Medical vision-language models (VLMs) transfer across label spaces through class names and prompt descriptions, either zero-shot or through parameter-efficient adaptation \citep{radford2021learning,wang2022medclip,zhang2023biomedclip,xie2025medtrinity,koleilat2025biomedcoop}. Prompt learning, medical knowledge mining, calibration-aware adaptation, and dynamic biomedical prompting have strengthened this interface \citep{zhou2022coop,liu2025kpl,basu2025calibprompt,koleilat2025biomedcoop,miao2026biodpp,shao2026vmfcoop,cui2026biomedccpl}. The resulting classifier depends jointly on visual geometry, textual class representations, prompt wording, and the cohort used for calibration.

Clinical shift tests every part of that dependency. Hospitals differ in acquisition protocol, hardware, patient population, disease prevalence, and annotation practice. Confidence learned on one cohort can remain high after the supporting visual geometry has changed \citep{ovadia2019can,xia2024cares,gutbrod2025openmibood,cheng2025robustness}. Average accuracy and marginal coverage offer incomplete summaries. A conformal predictor may cover nearly all patients in aggregate while repeatedly excluding the true label for one rare, visually unstable, or poorly supported disease class. Such failures are consequential precisely because the cohort-level metric appears satisfactory.

Several recent conformal directions bear directly on this problem. Localized conformal prediction for VLMs (LCP-VLM) adapts calibration to the neighborhood of a test image \citep{guan2023localized,hore2025localweights,fuchs2026lcpvlm}. Conformal prediction with optimal transport (Conf-OT) operates over the calibration and query pools, while Laplacian-Assisted Transductive Adaptation (LATA) refines medical VLM scores on an image graph \citep{silvarodriguez2025confot,bozorgtabar2026lata}. Empirical-Bayes conformal prediction with $r$-values ($\mathrm{CP}_{r\text{-value}}$) uses repeated VLM evidence to model score instability \citep{zeng2026cpr}. Tail-Aware Conformal Prediction (TACP) and soft Tail-Aware Conformal Prediction (sTACP) rebalance coverage using source-label frequency \citep{liu2026tacp}. These methods address local heterogeneity, set efficiency, prompt variability, or long-tailed labels. External clinical shift introduces a different class-tail problem: the least reliable target class need not be a source-frequency tail, and its identity can change with the backbone geometry. A fixed class partition or a marginal local threshold can therefore leave the target-limiting class exposed.

To address this failure, we introduce Class-Tail Adaptive Localized Conformal Deferral (CALCoDe), which separates class-tail discovery from independent protection. Out-of-fold validation predictions identify classes whose localized conformal behavior signals undercoverage risk. After this set of fragile classes is fixed, a disjoint calibration split supplies their class-conditional quantiles. A one-sided maximum combines each class-tail threshold with the localized threshold, adding protection for a fragile class while preserving every label admitted by the localized rule. Trimmed prompt evidence defines an APS nonconformity score, embedding-space localization adapts its threshold, and a separately calibrated support audit governs deferral when the representation lacks inlier support.

\begin{figure*}[t]
\centering
\begin{subfigure}[t]{0.515\textwidth}
  \centering
  \includegraphics[width=\linewidth]{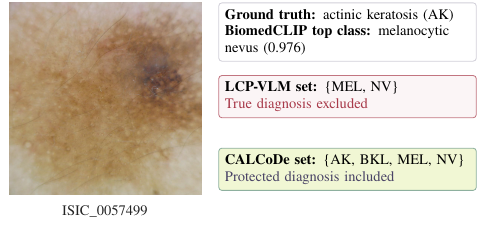}
  \caption{Accepted case with a missing true label.}
  \label{fig:calcode-case}
\end{subfigure}\hfill
\begin{subfigure}[t]{0.465\textwidth}
  \centering
  \includegraphics[width=\linewidth]{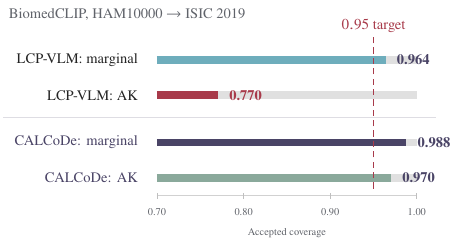}
  \caption{Marginal coverage conceals class-tail undercoverage.}
  \label{fig:calcode-coverage}
\end{subfigure}
\caption{Marginal accepted coverage masks class-tail undercoverage for BiomedCLIP on HAM10000$\rightarrow$ISIC 2019. (a) LCP-VLM excludes the histopathology-confirmed AK label for ISIC\_0057499; CALCoDe includes it. (b) LCP-VLM reaches 0.964 marginal accepted coverage but only 0.770 accepted coverage for AK; CALCoDe raises AK coverage to 0.970. Abbreviations: AK (actinic keratosis); BKL (benign keratosis-like lesion); MEL (melanoma); NV (melanocytic nevus). Image source: ISIC Archive, ISIC\_0057499, Hospital Cl\'inic de Barcelona; CC BY-NC.}
\label{fig:calcode-motivation}
\end{figure*}

Figure~\ref{fig:calcode-motivation} makes the class-tail failure concrete. For an accepted ISIC 2019 case with histopathology-confirmed actinic keratosis, BiomedCLIP assigns its largest probability to melanocytic nevus. LCP-VLM returns a two-label set that excludes the true diagnosis. The cohort-level statistic appears satisfactory--accepted coverage is 0.964--yet coverage for actinic keratosis is only 0.770. CALCoDe includes the protected diagnosis in this case and raises its class coverage to 0.970. Across HAM10000$\rightarrow$ISIC 2019 and HAM10000$\rightarrow$PAD-UFES-20, its observed marginal and worst-class accepted coverage exceed 0.95 in all eight backbone--dataset settings. These external-site values are empirical; the finite-sample result concerns protected classes exchangeable with the accepted calibration data.

\section{Contributions}

Our work makes three contributions to the reliability of medical VLM classification under clinical shift. First, we turn worst-class undercoverage from a post-hoc diagnostic into an adaptive calibration target: vulnerable classes are identified from out-of-fold coverage behavior rather than source prevalence and protected using independently estimated class-conditional thresholds. The resulting one-sided construction contains the corresponding localized conformal set and yields finite-sample coverage for protected classes under accepted-example exchangeability. Second, CALCoDe defines a leakage-controlled selective inference protocol with separate data roles for representation-support estimation, design selection, audit calibration, and conformal calibration. The protocol combines robust image--text evidence with a calibrated three-action interface while keeping the medical VLM unchanged. Third, we evaluate class-tail reliability across two external dermatology shifts and four frozen VLM backbones against standard and recent conformal methods. Class-wise coverage, paired bootstrap intervals, full-set frequency, and autonomous informative rate distinguish class-tail protection from coverage achieved through nearly complete prediction sets, thereby identifying regimes in which frozen representation geometry limits clinically useful automation.

\section{Problem Statement}

Let $\mathcal{X}$ denote the space of medical images and let $\mathcal{Y}=\{1,\ldots,K\}$ denote a fixed set of diagnostic classes for a target classification task. A medical vision-language classifier consists of an image encoder $f_{\theta}:\mathcal{X}\rightarrow \mathcal{Z}$, a set of text representations for class descriptions or prompts, and a scoring function that assigns class evidence to each label. For an input image $x$, the classifier produces a vector of class scores or probabilities $p(y \mid x)$ over $\mathcal{Y}$.

The conventional classification problem returns a single label
\[
\hat{y}(x)=\arg\max_{y\in\mathcal{Y}} p(y \mid x).
\]
This forced-decision setting is poorly matched to medical deployment. If the image is ambiguous, affected by acquisition artifacts, drawn from a shifted domain, or visually consistent with conditions outside the label space, a single top-1 prediction can be misleading even when the classifier assigns high confidence. Set-valued classification with abstention has a natural statistical interpretation in this setting because an empty set can indicate that no class is sufficiently supported by the reference distribution \citep{guan2022bcops}. The research problem is to construct a decision function
\begin{equation}
g(x)\in \mathcal{Y}\cup \{C:C\subseteq\mathcal{Y}, |C|>1\}\cup \{\mathrm{defer}\},
\label{eq:decision-space}
\end{equation}
where $g(x)$ may return a single label, a non-singleton set of plausible labels, or a deferral decision. Empty prediction sets are treated as deferrals because they indicate that no candidate label satisfies the calibrated reliability criterion.

The desired behavior is constrained by three requirements. First, when the input is consistent with the calibration distribution and the classifier has strong evidence, the system should return a compact decision, ideally a single label. Second, when multiple labels remain plausible, the system should return a prediction set whose empirical coverage is controlled at a user-specified risk level. Third, when the input is sufficiently shifted from the reliable operating regime, the system should defer in place of a falsely precise diagnosis. The main design challenge is to maintain safety without making the system clinically unusable: excessive deferral reduces utility, while overly small prediction sets can conceal uncertainty.

We formalize this tradeoff through a selective decision risk. For non-deferred examples, let $C_g(x)\subseteq\mathcal{Y}$ denote the returned singleton or prediction set; for deferred examples, define $C_g(x)=\emptyset$. Let the nonnegative weights $c_{\mathrm{err}}$, $c_{\mathrm{set}}$, and $c_{\mathrm{def}}$ encode the deployment costs of an uncovered true label, an overly broad diagnostic set, and a deferral, respectively. A cost-sensitive selective-risk objective is
\begin{equation}
R(g)=\mathrm{E}\big[c_{\mathrm{err}}L_{\mathrm{err}}(X,Y)+c_{\mathrm{set}}L_{\mathrm{set}}(X)+c_{\mathrm{def}}L_{\mathrm{def}}(X)\big],
\label{eq:selective-risk}
\end{equation}
where $L_{\mathrm{err}}=\mathbf{1}\{Y\notin C_g(X),g(X)\neq\mathrm{defer}\}$, $L_{\mathrm{set}}=|C_g(X)|$, and $L_{\mathrm{def}}=\mathbf{1}\{g(X)=\mathrm{defer}\}$,
subject to a target coverage constraint on accepted in-regime examples,
\begin{equation}
\Pr\{Y\in C_g(X)\mid g(X)\neq \mathrm{defer}, X\in\mathcal{R}\}\geq 1-\delta,
\label{eq:accepted-coverage}
\end{equation}
where $\mathcal{R}$ denotes the intended operating region represented by the calibration distribution. The costs express clinical preference and are specified outside the algorithm. For example, a screening tool may tolerate larger sets to avoid missed findings, whereas a triage tool may prefer deferral when evidence is weak.

The difficulty is that medical distribution shift affects both uncertainty estimation and coverage. Standard confidence scores are often miscalibrated under shift \citep{guo2017calibration,ovadia2019can}. Out-of-distribution (OOD) scores can detect some distributional deviations but may confuse challenging in-distribution examples with OOD inputs \citep{gutbrod2025openmibood}. Conformal prediction provides finite-sample coverage statements under exchangeability, yet those statements fail to extend automatically to shifted test distributions \citep{tibshirani2019conformal}. Medical VLMs add another source of instability: predictions depend jointly on the image representation, textual prompts, class descriptions, and image-text alignment margins \citep{zhou2022coop,koleilat2025biomedcoop,mahapatra2026valiant}.

We consider a post-hoc deployment setting with a pretrained or adapted medical VLM and labeled data from its intended operating distribution. Reference data support embedding-space comparisons, validation data govern score design, a gate-calibration split calibrates inlier support, and a conformal-calibration split constructs prediction sets. The objective is a decision rule that reduces high-confidence errors and weak-class undercoverage under clinically plausible shifts without backbone retraining, hidden training data, test labels, or complete metadata for every source of shift. Cross-fitting can implement these roles when labeled data are scarce, provided that score tuning, gate calibration, and prediction-set calibration retain separate label information.

\section{Proposed Method}

\subsection{Overview}

Class-Tail Adaptive Localized Conformal Deferral (CALCoDe) wraps a frozen medical VLM with a selective set-valued decision rule. Its core operation is adaptive class-tail protection: a localized conformal set remains the default, while classes that fail an out-of-fold validation coverage criterion receive independently calibrated class-conditional thresholds. Trimmed prompt probabilities supply an Adaptive Prediction Sets (APS) nonconformity score, and embedding-space neighborhoods localize its calibration. A separate support audit determines whether an input enters prediction-set construction.

CALCoDe assigns five distinct roles to the available data. A reference set $\mathcal{D}_{\mathrm{ref}}$ supplies embeddings for the support audit. A validation set $\mathcal{D}_{\mathrm{val}}$ selects the audit diagnostic and supplies out-of-fold class-tail discovery. A gate-calibration set $\mathcal{D}_{\mathrm{gate}}$, drawn from the intended inlier distribution, calibrates the deferral score after diagnostic selection. A conformal-calibration set $\mathcal{D}_{\mathrm{cal}}$ is accessed only after the audit and protected class set are frozen; it calibrates the localized and class-conditional thresholds. The test cohort $\mathcal{D}_{\mathrm{test}}$ is reserved for evaluation. This ordering prevents final-calibration scores from selecting the classes they subsequently protect.

For an input $x$, CALCoDe first aggregates prompt-level class evidence and converts it to a label-wise APS score. The audit either defers the image or passes it to embedding-localized calibration. The class-tail guard is then applied only to validation-identified fragile classes, after which the resulting set is returned as a singleton or a non-singleton prediction set.

\subsection{Robust Prompt Evidence}

The first stage stabilizes image--text class evidence against prompt wording. For each class $y$, let $\mathcal{T}_y=\{t_{y,1},\ldots,t_{y,M}\}$ denote clinically meaningful prompt variants. Let $z=f_{\theta}(x)$ be the image embedding and $e_{y,m}$ be the text embedding of prompt $t_{y,m}$. We use $\operatorname{sim}(\cdot,\cdot)$ for cosine similarity between normalized image and text embeddings, and $T>0$ for the temperature parameter that scales the similarity logits. For prompt index $m$, the VLM induces
\begin{equation}
\begin{aligned}
\ell_m(x,y)&=\frac{\operatorname{sim}(z,e_{y,m})}{T},\\
p_m(y\mid x)&=
\frac{\exp(\ell_m(x,y))}
{\sum_{k=1}^{K}\exp(\ell_m(x,k))}.
\end{aligned}
\label{eq:prompt-probability}
\end{equation}
Prompt ensembling is useful but fragile: an isolated prompt can distort class evidence. CALCoDe uses trimmed prompt aggregation. Let $\mathcal{M}_{\kappa}(x,y)$ be the prompt indices remaining after removing the $\kappa$ largest and $\kappa$ smallest values of $\{p_m(y\mid x)\}_{m=1}^{M}$. Define
\begin{equation}
\begin{aligned}
a_y(x)&=\frac{1}{|\mathcal{M}_{\kappa}(x,y)|}
\sum_{m\in\mathcal{M}_{\kappa}(x,y)}p_m(y\mid x),\\
p_y^{\mathrm{rob}}(x)&=
\frac{a_y(x)}{\sum_{k=1}^{K}a_k(x)+\varepsilon}.
\end{aligned}
\label{eq:robust-prompt-evidence}
\end{equation}
Trimmed aggregation preserves the post-hoc interface and reduces sensitivity to isolated prompt failures.

\subsection{APS Nonconformity Score}

The trimmed probabilities in Eq.~\eqref{eq:robust-prompt-evidence} are converted to a rank-sensitive APS score \citep{angelopoulos2021uncertainty}. Let $\pi_x(1),\ldots,\pi_x(K)$ order the labels by decreasing $p_y^{\mathrm{rob}}(x)$, and let $r_x(y)$ be the rank of candidate label $y$. CALCoDe uses
\begin{equation}
s_{\mathrm{APS}}(x,y)=
\sum_{j=1}^{r_x(y)}p_{\pi_x(j)}^{\mathrm{rob}}(x).
\label{eq:label-nonconformity}
\end{equation}
The score is small for labels near the head of the robust prompt distribution and increases with the cumulative mass preceding a candidate label. Eq.~\eqref{eq:label-nonconformity} supplies the nonconformity values for localized calibration, out-of-fold tail discovery, and the final class-tail guard.

\subsection{Audited Conformal Deferral}

The label-wise score in Eq.~\eqref{eq:label-nonconformity} governs set membership, whereas deferral requires a sample-level assessment. CALCoDe forms candidate unreliability scores $B_a(x)$, with larger values indicating weaker support. The base candidates are robust global distance, negative energy, negative maximum robust prompt probability, and prompt-distribution disagreement. The global support score is
\[
D_{\mathrm{glob}}^{\mathrm{rob}}(x)
=\operatorname{median}\{d_{(1)}(z),\ldots,d_{(k)}(z)\},
\]
where $d_{(j)}(z)$ is the $j^{\mathrm{th}}$ smallest distance from $z$ to the pooled inlier reference embeddings. Prompt disagreement is measured by the average divergence between prompt-specific class distributions and the trimmed aggregate distribution:
\[
B_{\mathrm{prompt}}(x)=
\frac{1}{M}\sum_{m=1}^{M}
\mathrm{KL}\!\left(p_m(\cdot\mid x)\,\|\,p^{\mathrm{rob}}(\cdot\mid x)\right).
\]
Here $\mathrm{KL}(u\,\|\,v)$ denotes Kullback--Leibler divergence from distribution $u$ to distribution $v$.

For each base diagnostic, its empirical gate rank yields a support value $\widetilde p_a(x)$. A fused-rank candidate,
\begin{equation}
B_{\mathrm{fuse}}(x)=1-\min_a\widetilde p_a(x),
\label{eq:fused-rank-audit}
\end{equation}
responds when any base diagnostic assigns weak inlier support. The candidate set for audit selection contains the four base diagnostics and $B_{\mathrm{fuse}}$.

The validation split determines the diagnostic. Let $\hat{y}(x)=\arg\max_y p_y^{\mathrm{rob}}(x)$ and define $q_i=\mathbf{1}\{\hat{y}(x_i)\neq y_i\}$ for validation examples. CALCoDe selects
\begin{equation}
a^\star=\arg\max_a \mathrm{AUROC}\big(B_a(x_i),q_i\big),
\label{eq:audit-selection}
\end{equation}
where AUROC denotes area under the receiver operating characteristic curve. CALCoDe then fixes $B_{a^\star}$ before accessing the gate-calibration split. For a test image $x$, the audited inlier p-value is
\begin{equation}
p_{\mathrm{audit}}(x)=
\frac{1+\sum_{j=1}^{n_g}\mathbf{1}\{B_{a^\star}(x_j)\geq B_{a^\star}(x)\}}
{n_g+1},
\label{eq:audit-pvalue}
\end{equation}
where $x_j\in\mathcal{D}_{\mathrm{gate}}$. The acceptance indicator is $A(x)=\mathbf{1}\{p_{\mathrm{audit}}(x)\geq\alpha_{\mathrm{def}}\}$. For a selected base diagnostic fixed before gate calibration, Supplementary Section S1.1 provides the usual conformal rank statement under inlier exchangeability. The fused-rank candidate in Eq.~\eqref{eq:fused-rank-audit} uses empirical gate ranks and is treated as a calibrated relative-support diagnostic rather than as a separate distribution-free test. Severe localized covariate shift also violates gate--target exchangeability and can loosen false-alarm control. Target-cohort deferral is therefore reported empirically and carries no distribution-free OOD guarantee.

\subsection{Localized Conformal Calibration}

Inputs accepted by Eq.~\eqref{eq:audit-pvalue} proceed to localized prediction-set calibration. The accepted calibration subset is
\begin{equation}
\mathcal{D}_{\mathrm{cal}}^{A}
=\{(x_i,y_i)\in\mathcal{D}_{\mathrm{cal}}:
p_{\mathrm{audit}}(x_i)\geq\alpha_{\mathrm{def}}\}.
\label{eq:accepted-calibration}
\end{equation}
For each accepted calibration example, compute $r_i=s_{\mathrm{APS}}(x_i,y_i)$ using Eq.~\eqref{eq:label-nonconformity}. A global split-conformal threshold weights all accepted calibration scores equally, although acquisition and phenotype can change the relevance of those examples to a particular image. We therefore localize calibration in the VLM embedding space \citep{guan2023localized,hore2025localweights}. For an accepted test image $x$, define
\begin{equation}
\begin{aligned}
H_h(x,x_i)&=\exp\{-d(z,z_i)^2/h^2\},\\
w_i(x)&=\frac{H_h(x,x_i)}
{1+\sum_j H_h(x,x_j)}.
\end{aligned}
\label{eq:localized-weights}
\end{equation}
The remaining mass $1/(1+\sum_jH_h(x,x_j))$ is placed at $\infty$, yielding the conservative weighted empirical distribution used in localized conformal prediction. The local quantile level $\eta_{\delta,h}$ is chosen by leave-one-out calibration on $\mathcal{D}_{\mathrm{cal}}^{A}$ so that the accepted calibration scores are covered at target level $1-\delta$. The resulting threshold $\tau_{\delta,h}(x)$ defines
\begin{equation}
C_{\delta,h}^{\mathrm{loc}}(x)=
\{y\in\mathcal{Y}:s_{\mathrm{APS}}(x,y)\leq\tau_{\delta,h}(x)\}.
\label{eq:localized-set}
\end{equation}
When localization is disabled, the construction reduces to ordinary split conformal prediction on accepted examples. Supplementary Section S1.2 states the corresponding accepted-set exchangeability condition.

\subsection{Validation-Discovered Class-Tail Protection}

Localized calibration adapts to the neighborhood of a test image, yet Eq.~\eqref{eq:localized-set} can cover most accepted examples while leaving a rare or unstable class under-covered. CALCoDe identifies such classes before the final calibration split is accessed. Let
\[
V_y^A=\{i:(x_i,y_i)\in\mathcal{D}_{\mathrm{val}},\,
y_i=y,\,p_{\mathrm{audit}}(x_i)\geq\alpha_{\mathrm{def}}\}.
\]
Partition the accepted validation examples into $F$ stratified folds. For fold $f$, fit a pilot localized rule $C_{\delta,h}^{(-f)}$ using the remaining $F-1$ folds and evaluate only the held-out examples. CALCoDe estimates class-wise accepted coverage from these out-of-fold decisions,
\begin{equation}
\widehat{\mathrm{Cov}}_y^{\mathrm{val}}
=
\frac{1}{|V_y^A|\vee 1}
\sum_{i\in V_y^A}
\mathbf{1}\{y_i\in C_{\delta,h}^{(-f(i))}(x_i)\}.
\label{eq:validation-class-coverage}
\end{equation}
Classes whose out-of-fold coverage falls below a margin-adjusted target are marked as tail risks. Classes with fewer than $n_{\min}$ accepted validation examples are protected by default because the available evidence cannot establish stability:
\begin{equation}
\begin{aligned}
\mathcal{Y}_{\mathrm{tail}}
={}&\{y:n_y^{\mathrm{val}}<n_{\min}\}\\
&\cup\{y:n_y^{\mathrm{val}}\geq n_{\min},\;
\widehat{\mathrm{Cov}}_y^{\mathrm{val}}<1-\delta-\gamma\}.
\end{aligned}
\label{eq:tail-risk-classes}
\end{equation}
The set $\mathcal{Y}_{\mathrm{tail}}$ is frozen before $\mathcal{D}_{\mathrm{cal}}$ is used. For each $y\in\mathcal{Y}_{\mathrm{tail}}$, CALCoDe computes the finite-sample-corrected class-global quantile at guard level $1-\delta_y$ from accepted calibration examples with label $y$. A class-local quantile can enlarge this value when enough class-$y$ examples are available. Their maximum defines $\tau_y^{\mathrm{tail}}$; an unavailable finite-sample quantile is represented by $+\infty$. The final class-specific threshold is
\begin{equation}
\tau_{\mathrm{CALCoDe}}(x,y)=
\begin{cases}
\max\{\tau_{\delta,h}(x),\tau_y^{\mathrm{tail}}\}, & y\in\mathcal{Y}_{\mathrm{tail}},\\
\tau_{\delta,h}(x), & y\notin\mathcal{Y}_{\mathrm{tail}}.
\end{cases}
\label{eq:calcode-threshold}
\end{equation}
The CALCoDe prediction set is
\begin{equation}
C_{\mathrm{CALCoDe}}(x)=
\{y\in\mathcal{Y}:s_{\mathrm{APS}}(x,y)\leq\tau_{\mathrm{CALCoDe}}(x,y)\}.
\label{eq:calcode-set}
\end{equation}
The maximum in Eq.~\eqref{eq:calcode-threshold} is the defining structural choice. It allocates additional calibration slack to a validation-identified fragile class while preserving every inclusion made by the localized rule.

\subsection{Theoretical Properties}

\begin{proposition}[Set monotonicity]
\label{prop:set-monotonicity}
For any $x$ and protected class
$y\in\mathcal{Y}_{\mathrm{tail}}$,
\begin{equation}
y\in C_{\delta,h}^{\mathrm{loc}}(x)
\quad\Longrightarrow\quad
y\in C_{\mathrm{CALCoDe}}(x).
\label{eq:main-tail-monotonicity}
\end{equation}
\end{proposition}

The result follows directly from
$\tau_{\mathrm{CALCoDe}}(x,y)\geq\tau_{\delta,h}(x)$.

\begin{proposition}[Protected-class coverage]
\label{prop:protected-class-coverage}
Fix $y\in\mathcal{Y}_{\mathrm{tail}}$ before accessing
$\mathcal{D}_{\mathrm{cal}}$. Suppose that the accepted class-$y$
calibration scores and an accepted class-$y$ test score are
exchangeable, and that $\tau_y^{\mathrm{tail}}$ contains the corrected
class-global quantile at level $1-\delta_y$. Then
\begin{equation}
\Pr\!\left\{
Y\in C_{\mathrm{CALCoDe}}(X)
\mid Y=y,\,A(X)=1
\right\}
\geq 1-\delta_y .
\label{eq:main-protected-coverage}
\end{equation}
\end{proposition}
Cross-fitted discovery and the disjoint final calibration split keep the class-selection event independent of the order statistic used in Eq.~\eqref{eq:main-protected-coverage}. Supplementary Sections S1.3 and S1.4 provide the discovery argument, both proofs, and the finite-sample correction. The experiments set $1-\delta_y=0.95$, matching the pooled target; the validation discovery boundary is 0.90. Observed target-site coverage remains empirical because target exchangeability is not assumed.

\subsection{Three-Action Decision Rule}

The calibrated audit p-value and guarded prediction set jointly determine the clinical action. Let $C_*(x)=C_{\mathrm{CALCoDe}}(x)$. The resulting three-action decision rule is
\begin{equation}
g(x)=
\begin{cases}
\mathrm{defer}, & p_{\mathrm{audit}}(x)<\alpha_{\mathrm{def}},\\
\mathrm{defer}, & |C_*(x)|=0,\\
y, & C_*(x)=\{y\},\\
C_*(x), & |C_*(x)|>1.
\end{cases}
\label{eq:three-action-rule}
\end{equation}
The three outputs correspond to distinct clinical actions: automated narrow prediction, bounded diagnostic ambiguity, and escalation outside the automated classifier. Empty sets are escalated because no label satisfies the calibrated criterion. For multiclass tasks, non-singleton sets can still narrow the plausible differential diagnosis, and set size is reported as a utility cost.

For cost-aware deployment, validation data can select an operating point that minimizes the selective risk in Eq.~\eqref{eq:selective-risk}, subject to accepted-coverage and informative-decision constraints. The selected policy is fixed before held-out reporting and adds no distributional guarantee.

Supplementary Section S2 positions CALCoDe relative to recent medical VLM adaptation, conformal calibration, class-tail protection, and OOD detection approaches.

\section{Experiments}

\subsection{Evaluation Setup}

The primary evaluation uses two external-shift dermatology settings. HAM10000 supplies the reference, validation, gate-calibration, and conformal-calibration splits. ISIC 2019 and PAD-UFES-20 serve as external test cohorts and differ from HAM10000 in acquisition protocol, population, prevalence, and image source. Four frozen image--text backbones span general and medical pretraining: BiomedCLIP \citep{zhang2023biomedclip}, OpenAI CLIP ViT-B/32 \citep{radford2021learning}, PubMedCLIP ViT-B/32 \citep{eslami2023pubmedclip}, and MedSigLIP-448 \citep{google2025medsiglip}. All methods receive the same cached model outputs, class names, prompt templates, and data partitions for a given backbone. Supplementary Section S3.1 provides the complete construction and hyperparameter values.

The main comparison includes adaptive prediction sets (APS) \citep{angelopoulos2021uncertainty}, Mondrian conformal prediction (Mondrian CP) \citep{vovk2005algorithmic}, localized conformal prediction (local CP) \citep{guan2023localized}, Conf-OT \citep{silvarodriguez2025confot}, LCP-VLM \citep{fuchs2026lcpvlm}, matched-prompt $\mathrm{CP}_{r\text{-value}}$ \citep{zeng2026cpr}, the APS variants of TACP and sTACP \citep{liu2026tacp}, and LATA \citep{bozorgtabar2026lata}. The complete LATA system includes a separately trained ViLU head. We therefore evaluate its graph-refinement component and label the comparison explicitly as ``LATA graph refinement (no ViLU).'' Supplementary Section S3.5 records each adaptation and reports the broader catalog of singleton, calibration, deferral, split-conformal, and regularized adaptive prediction sets (RAPS).

The fixed comparison point uses target accepted coverage $1-\delta=0.95$, class-tail guard level $1-\delta_y=0.95$, validation discovery boundary $1-\delta-\gamma=0.90$, and $\alpha_{\mathrm{def}}=0.05$. Every diagnostic, protected class set, threshold, and operating point is fixed before test evaluation. Coverage, worst-class coverage, and set size are computed on accepted examples; worst-class coverage is the minimum accepted coverage across disease classes. Deferral is measured over the full test cohort. Supplementary Section S3.3 defines the class-stratified paired bootstrap. Supplementary Section S4 reports the binary histopathology stress test, while Supplementary Section S5 reports full-set frequency, autonomous informative rate, and cost-sensitive policies.

The experiments address three questions. \textbf{Q1} asks whether CALCoDe protects the least-covered class under external shift; \textbf{Q2} examines how frozen-backbone geometry changes the support audit; and \textbf{Q3} identifies which components drive class-tail reliability.

\subsection{Q1: Does CALCoDe Protect the Least-Covered Class?}

Table~\ref{tab:external-shift-complete} reports accepted coverage, worst-class accepted coverage, average accepted set size, and deferral together. The first two quantities assess reliability; the latter two show how much utility is spent to obtain it. Reporting all four is necessary because a nearly complete label set can achieve excellent coverage while contributing limited diagnostic specificity.

\begin{table*}[t]
\centering
\scriptsize
\renewcommand{\arraystretch}{0.93}
\setlength{\tabcolsep}{2.4pt}
\textbf{(a) HAM10000$\rightarrow$ISIC 2019}\\[2pt]
\resizebox{\textwidth}{!}{%
\begin{tabular}{@{}lccccc@{}}
\hline
Method & BiomedCLIP & OpenAI CLIP & PubMedCLIP & MedSigLIP-448 & Average \\
\hline
APS & 0.912/0.195/5.23/-- & 1.000/1.000/7.00/-- & 0.850/0.093/6.06/-- & 0.947/0.473/5.52/-- & 0.927/0.440/5.95/-- \\
Mondrian CP & 0.947/0.928/5.90/-- & 0.921/0.747/6.39/-- & 0.944/0.769/6.32/-- & 0.938/0.791/5.70/-- & 0.938/0.809/6.08/-- \\
Local CP & 0.955/0.741/5.75/-- & 0.983/0.958/6.96/-- & 0.975/0.869/6.65/-- & 0.978/0.857/6.26/-- & 0.972/0.856/6.41/-- \\
Conf-OT APS & 0.916/0.593/4.76/-- & 0.847/0.055/3.97/-- & 0.921/0.077/4.60/-- & 0.926/0.022/4.37/-- & 0.903/0.187/4.43/-- \\
  $\mathrm{CP}_{r\text{-value}}$ (matched prompts) & 0.886/0.163/4.49/-- & 0.974/0.936/6.95/-- & 0.930/0.565/6.55/-- & 0.870/0.352/4.29/-- & 0.915/0.504/5.57/-- \\
TACP-APS & 0.926/0.211/5.64/-- & 1.000/1.000/7.00/-- & 0.970/0.927/6.92/-- & 0.970/0.923/6.64/-- & 0.966/0.765/6.55/-- \\
sTACP-APS & 0.951/0.853/6.39/-- & 1.000/1.000/7.00/-- & 0.960/0.927/6.85/-- & 0.967/0.925/6.70/-- & 0.970/0.926/6.74/-- \\
LATA graph refinement (no ViLU) & 0.936/0.204/5.68/-- & 0.999/0.998/7.00/-- & 0.888/0.325/6.30/-- & 0.976/0.714/5.98/-- & 0.950/0.560/6.24/-- \\
LCP-VLM APS & 0.964/0.770/5.95/-- & 0.981/0.956/6.94/-- & 0.969/0.862/6.61/-- & 0.978/0.868/6.30/-- & 0.973/0.864/6.45/-- \\
\hline
\rowcolor{CalCoDE-Row}\textbf{CALCoDe} & 0.988/0.970/6.27/0.203 & 0.980/0.954/6.94/0.072 & 0.997/0.996/6.89/0.157 & 0.992/0.960/6.76/0.095 & 0.989/0.970/6.71/0.132 \\
\hline
\end{tabular}}

\vspace{4pt}
\textbf{(b) HAM10000$\rightarrow$PAD-UFES-20}\\[2pt]
\resizebox{\textwidth}{!}{%
\begin{tabular}{@{}lccccc@{}}
\hline
Method & BiomedCLIP & OpenAI CLIP & PubMedCLIP & MedSigLIP-448 & Average \\
\hline
APS & 0.810/0.619/3.58/-- & 0.984/0.939/4.88/-- & 0.722/0.314/4.25/-- & 0.972/0.921/4.89/-- & 0.872/0.698/4.40/-- \\
Mondrian CP & 0.943/0.689/3.87/-- & 0.938/0.865/4.45/-- & 0.953/0.652/4.30/-- & 0.954/0.799/3.94/-- & 0.947/0.751/4.14/-- \\
Local CP & 1.000/0.996/4.99/-- & 1.000/0.996/5.00/-- & 0.990/0.975/4.98/-- & 0.998/0.988/4.99/-- & 0.997/0.989/4.99/-- \\
Conf-OT APS & 0.856/0.749/3.62/-- & 0.778/0.453/3.87/-- & 0.674/0.300/4.00/-- & 0.821/0.526/3.83/-- & 0.782/0.507/3.83/-- \\
  $\mathrm{CP}_{r\text{-value}}$ (matched prompts) & 0.801/0.587/3.42/-- & 0.982/0.852/4.93/-- & 0.820/0.553/4.60/-- & 0.655/0.141/3.00/-- & 0.814/0.533/3.99/-- \\
TACP-APS & 0.889/0.738/4.01/-- & 1.000/1.000/5.00/-- & 0.974/0.775/4.76/-- & 0.993/0.943/4.94/-- & 0.964/0.864/4.68/-- \\
sTACP-APS & 0.830/0.660/3.65/-- & 0.993/0.939/4.90/-- & 0.968/0.742/4.62/-- & 0.993/0.943/4.94/-- & 0.946/0.821/4.53/-- \\
LATA graph refinement (no ViLU) & 0.859/0.666/3.98/-- & 1.000/1.000/5.00/-- & 0.750/0.383/4.31/-- & 0.961/0.889/4.83/-- & 0.893/0.735/4.53/-- \\
LCP-VLM APS & 1.000/0.981/4.99/-- & 1.000/1.000/4.99/-- & 0.994/0.985/4.98/-- & 0.999/0.992/4.99/-- & 0.998/0.989/4.99/-- \\
\hline
\rowcolor{CalCoDE-Row}\textbf{CALCoDe} & 0.997/0.960/4.98/0.832 & 1.000/1.000/5.00/0.009 & 1.000/1.000/5.00/0.403 & 1.000/0.996/5.00/0.042 & 0.999/0.989/4.99/0.321 \\
\hline
\end{tabular}}
\caption{Class-tail reliability across two external dermatology shifts and four frozen VLM backbones. Each cell reports marginal accepted coverage / worst-class accepted coverage / mean accepted-set size / deferral rate. The final column averages each metric across backbones; ``--'' denotes methods without an explicit deferral mechanism.}
\label{tab:external-shift-complete}
\end{table*}
On ISIC 2019, CALCoDe reaches 0.989 accepted coverage and 0.970 worst-class coverage averaged over four backbones. The corresponding values are 0.970 and 0.926 for sTACP, and 0.973 and 0.864 for LCP-VLM. CALCoDe's average accepted set contains 6.71 of seven labels, improving slightly upon sTACP's 6.74 while remaining close to LCP-VLM's 6.45; its deferral rate is 0.132. OpenAI CLIP clarifies why coverage cannot be read alone: APS and TACP reach 1.000 by returning essentially the full seven-class set. Supplementary Section S5 reports full-set frequency and autonomous informative rate, separating conservative saturation from useful specificity.

The paired class-stratified intervals in Supplementary Figure S1 and Table S2 show that CALCoDe improves accepted and worst-class coverage over sTACP on both shifts and over LCP-VLM on ISIC 2019; the PAD-UFES-20 differences from LCP-VLM include zero.

PAD-UFES-20 marks the utility boundary. CALCoDe averages 0.999 accepted coverage and 0.989 worst-class coverage, while LCP-VLM obtains 0.998 and 0.989. Their paired differences include zero. Both methods return approximately 4.99 labels from a five-class space; CALCoDe additionally defers 0.321 of cases on average. Its reliability improvement over sTACP is significant, but the resulting outputs remain largely non-specific. We treat this cohort as evidence that post-hoc uncertainty control cannot recover class separation absent from the frozen representation.

Across all eight dataset--backbone settings, CALCoDe is the only evaluated method with observed marginal and worst-class accepted coverage both at or above 0.95 in every setting. LCP-VLM meets the marginal target in 8/8 settings and the worst-class target in 5/8; sTACP meets them in 7/8 and 1/8. These counts summarize empirical external-shift behavior rather than a target-domain guarantee. Class-wise sample sizes and Wilson intervals in Supplementary Section S3.3 qualify the point estimates for small limiting classes.

\subsection{Q2: How Does Backbone Geometry Change the Audit?}

Supplementary Section S3.2 reports the full PAD-UFES-20 geometry audit. BiomedCLIP has the largest class-local shift ratio, $5.80$, the lowest median audited p-value, $0.004$, and lower effective covariance rank than OpenAI CLIP, $38.53$ versus $47.99$. The lower effective rank quantitatively supports the representation-anisotropy and class-local over-clustering hypothesis under the shift from dermoscopy to clinical photography. OpenAI CLIP and MedSigLIP defer less often, but their accepted sets remain nearly full and their singleton rate is zero; deferral and diagnostic specificity must therefore be interpreted jointly.

\subsection{Q3: Which Components Drive Class-Tail Reliability?}

Table~\ref{tab:main-ablation} changes one component at a time within the verified APS implementation and averages results over both dermatology shifts and four backbones. Removing the class-tail guard lowers worst-class accepted coverage from 0.979 to 0.917, and removing localization lowers it to 0.900. Prompt trimming provides a smaller but consequential gain: without it, worst-class coverage is 0.970 and reaches 0.95 in only six of the eight settings. Guarding every class raises worst-class coverage to 0.997 but increases mean set size by 0.13 labels, showing that validation-based discovery concentrates the expansion on classes with observed coverage failure. Removing the support audit leaves accepted coverage nearly unchanged while reducing deferral to zero. The guard and localization therefore drive class-tail coverage, whereas the audit controls which cases enter the automated prediction-set pathway. Supplementary Section S3.4 reports per-setting ablations and target-attainment counts.

Increasing $\alpha_{\mathrm{def}}$ from 0.01 to 0.10 raises average deferral and selective accuracy, as expected for a stricter gate. We use $\alpha_{\mathrm{def}}=0.05$ as the fixed comparison point rather than selecting it on the test cohorts. Supplementary Section S5 instantiates the risk in Eq.~\eqref{eq:selective-risk} under four cost regimes and reports validation-selected operating points. When deferral is expensive, validation selection lowers mean held-out deferral by 0.150 across the eight dermatology settings for the non-tail-guarded policy. The near-full PAD-UFES-20 sets persist, confirming that operating-point selection cannot recover diagnostic specificity from an incompatible frozen representation.

\begin{table}[H]
\centering
\scriptsize
\setlength{\tabcolsep}{3.2pt}
\begin{tabular}{lcccc}
\hline
Variant & Cov.$\uparrow$ & Worst cov.$\uparrow$ & Set$\downarrow$ & Def.$\downarrow$ \\
\hline
Without class-tail guard & 0.985 & 0.917 & 5.68 & 0.227 \\
Without adaptive discovery & 0.998 & 0.997 & 5.98 & 0.226 \\
Without support audit & 0.995 & 0.983 & 5.87 & 0.000 \\
Without localization & 0.976 & 0.900 & 5.77 & 0.226 \\
Without prompt trimming & 0.992 & 0.970 & 5.85 & 0.226 \\
\hline
\rowcolor{CalCoDE-Row}\textbf{CALCoDe} & 0.994 & 0.979 & 5.85 & 0.226 \\
\hline
\end{tabular}
\caption{Contributions of CALCoDe components to class-tail reliability.}
\label{tab:main-ablation}
\end{table}

\section{Discussion}

\subsection{Class-Tail Failure and Representation Failure}

Clinical shift exposes reliability structure that marginal coverage can conceal. LCP-VLM preserves marginal coverage while leaving three ISIC 2019 backbones below the worst-class target, and frequency-based TACP variants transfer unevenly when source prevalence does not identify the limiting target class. CALCoDe uses held-out coverage behavior to identify fragile classes and reserves an independent split for their calibration. This data separation matters: a class enters the guard because its out-of-fold predictions reveal undercoverage, not because its test performance is known or its source count is small. The one-sided maximum then increases the threshold only where the validation evidence calls for protection, while retaining every label admitted by the localized conformal rule.

The experiments distinguish a correctable calibration failure from a representation-limited failure. On ISIC 2019, the frozen embeddings preserve enough class-relevant structure for targeted threshold expansion to raise the least-covered class without forcing every prediction to contain all seven labels. On PAD-UFES-20, the shift from dermoscopy to clinical photography weakens diagnostic separation, and both CALCoDe and LCP-VLM approach the complete five-class set. Coverage remains high in this regime, but the resulting sets carry limited diagnostic specificity. A post-hoc calibration layer can redistribute uncertainty already represented by the backbone; it cannot create class separation absent from the embedding space.

The geometry audit helps locate this boundary. BiomedCLIP exhibits a class-local shift ratio of 5.80, a median audited p-value of 0.004, and a lower effective covariance rank than OpenAI CLIP (38.53 versus 47.99) on PAD-UFES-20. The lower rank quantitatively supports the representation-anisotropy and class-local over-clustering explanation: a narrower source geometry leaves target clinical photographs with little local support. OpenAI CLIP and MedSigLIP trigger fewer deferrals, yet their accepted sets remain nearly full. Deferral frequency therefore reflects the interaction between the audit and the backbone geometry; it is not, by itself, a measure of diagnostic usefulness.

\subsection{Interpreting Prediction Sets and Deferral}

CALCoDe exposes three outcomes that have different operational meanings. A singleton supports autonomous classification at the selected operating point. A non-singleton set preserves plausible alternatives for downstream review. Deferral indicates insufficient calibrated support for either form of automated output. Their clinical value depends on the label space and workflow. A two-label set on a binary histopathology task communicates no diagnostic preference, whereas a two-label set in a seven-class dermatology task can still narrow the differential diagnosis. For this reason, coverage, worst-class coverage, set size, full-set frequency, singleton rate, and deferral should be reported together.

This joint view also prevents favorable interpretations based on one metric alone. Low deferral may coexist with nearly complete prediction sets, as observed for several PAD-UFES-20 backbones. Conversely, a higher deferral rate may be appropriate when an error on a fragile class carries a greater clinical cost than review. The cost analysis in Supplementary Section S5 makes this dependence explicit: the preferred operating point changes with the relative costs of error, set ambiguity, and deferral. CALCoDe supplies calibrated statistics and a three-action interface; the deployment institution must specify the cost regime and available review capacity.

\subsection{Statistical Scope}

The protected-class statement rests on two design conditions. The vulnerable classes are fixed using cross-fitted validation predictions before the calibration split is accessed, and the accepted calibration and test scores are exchangeable within each protected class. Under these conditions, the corrected class-conditional quantile yields finite-sample coverage at the guard level, while set monotonicity follows deterministically from the maximum operator. The latter property holds for every input and does not depend on exchangeability.

External clinical shift requires a narrower interpretation. Test examples from a new acquisition protocol need not be exchangeable with accepted calibration examples, so target-site coverage is an empirical result rather than a distribution-free guarantee. The gate p-value has its usual conformal type-I interpretation only for inlier examples exchangeable with the gate split. On a shifted cohort, it serves as a calibrated relative-support index; severe localized shift can make its false-alarm rate loose. The experiments consequently report external-site coverage, class-wise coverage, and audit behavior without extending the finite-sample statement to arbitrary covariate shift.

\subsection{Limitations and Research Directions}

The study centers on two external dermatology shifts, with Camelyon17-WILDS used as a histopathology stress test. Broader evidence is needed across radiology, ophthalmology, multimodal clinical records, and prospective multi-site workflows. The experiments also use frozen backbones and fixed label spaces. Representation adaptation may recover specificity in the saturation regime, although its training data and calibration data must remain separated to preserve the stated inference protocol. Hierarchical label spaces offer another path: a system could retain coarse diagnostic informativeness when fine-grained classes are not separable.

CALCoDe requires labeled reference, validation, gate, and calibration partitions. This sample-splitting discipline limits leakage but can reduce the number of examples available for rare-class calibration. Multi-site calibration, principled partial pooling across related classes, and sequential recalibration as reviewed labels accumulate are promising extensions. Such procedures would need explicit control of selection and temporal dependence. Prospective evaluation should further measure review time, downstream diagnostic decisions, and subgroup-specific consequences rather than treating deferral as an abstract scalar cost.

\section{Conclusion}

Marginal coverage can hide the disease class most exposed by clinical shift. CALCoDe makes that failure an explicit calibration target through cross-fitted class-tail discovery, independent class-conditional calibration, and a one-sided localized conformal guard. The construction contains the corresponding localized conformal set and provides finite-sample protected-class coverage under accepted-example exchangeability. Across eight external-shift settings, CALCoDe is the only evaluated method whose observed marginal and worst-class accepted coverage both reach the 0.95 target in every setting.

The results also identify the boundary of post-hoc uncertainty correction. When the frozen representation retains class-relevant support, targeted calibration can repair class-specific undercoverage while preserving the localized baseline. When representations collapse under a severe acquisition shift, prediction sets expand toward the full label space and cease to provide useful diagnostic discrimination. That behavior should be treated as evidence that the representation, label space, or clinical evidence requires revision, rather than as a successful deployment outcome based on coverage alone.

Class-tail coverage, prediction-set specificity, and deferral together provide a more informative reliability profile for medical VLMs than marginal coverage in isolation. CALCoDe offers a statistically disciplined way to construct that profile without retraining the underlying model. Its broader implication is methodological: reliability layers for clinical foundation models should identify which classes fail, state the assumptions under which protection holds, and reveal when calibrated uncertainty has reached the limit imposed by the representation.

\input{supplement.tex}

\bibliographystyle{plainnat}
\bibliography{references}

\end{document}

%% file: supplement.tex
\clearpage
\appendix
\renewcommand{\thesection}{S\arabic{section}}
\renewcommand{\thesubsection}{S\arabic{section}.\arabic{subsection}}
\renewcommand{\thetable}{S\arabic{table}}
\renewcommand{\thefigure}{S\arabic{figure}}
\setcounter{section}{0}
\setcounter{table}{0}
\setcounter{figure}{0}
\setcounter{secnumdepth}{2}

\begin{center}
{\LARGE\bfseries Supplementary Material}\\[0.5em]
{\large Mitigating Class-Tail Undercoverage in Medical Vision-Language Models under Clinical Shift}
\end{center}
\vspace{1em}

\section{Finite-Sample Statements}
\label{sec:supp-gate-validity}

\subsection{Audited Inlier-Support Rank Validity}
\label{sec:supp-audit-rank}

The audited gate uses validation data to choose a scalar reliability diagnostic and an independent gate split to calibrate that diagnostic. Sample splitting yields the following finite-sample statement. Suppose the selected diagnostic $B_{a^\star}$ is fixed without using $\mathcal{D}_{\mathrm{gate}}$, and let $X$ be an inlier test example exchangeable with the gate examples $\{X_j\}_{j=1}^{n_g}$. Define
\begin{equation}
p_{\mathrm{audit}}(X)=
\frac{1+\sum_{j=1}^{n_g}\mathbf{1}\{B_{a^\star}(X_j)\geq B_{a^\star}(X)\}}
{n_g+1}.
\label{eq:supp-audit-pvalue}
\end{equation}
Then, for any $\alpha\in[0,1]$,
\begin{equation}
\Pr\{p_{\mathrm{audit}}(X)\leq \alpha\}\leq \alpha .
\label{eq:supp-superuniform}
\end{equation}
The proof is the standard conformal rank argument: conditional on the multiset of gate and test scores, the rank of the exchangeable test score among the $n_g+1$ scores is uniform up to ties, and the conservative numerator with the added one makes the resulting p-value super-uniform. This statement applies when the selected scalar diagnostic is fixed independently of the gate split. CALCoDe also considers a fused-rank diagnostic constructed from empirical gate ranks; that candidate is interpreted as a calibrated relative-support score and is not covered by the fixed-score rank statement. For a deployment cohort that is not exchangeable with $\mathcal{D}_{\mathrm{gate}}$, either audit score remains an empirical inlier-support diagnostic without distribution-free false-alarm control.

\subsection{Accepted-Set Exchangeability}
\label{sec:supp-accepted-exchangeability}

After the gate is fixed, accepted calibration examples are
\begin{equation}
\mathcal{D}_{\mathrm{cal}}^{A}=\{(X_i,Y_i):p_{\mathrm{audit}}(X_i)\geq \alpha_{\mathrm{def}}\}.
\label{eq:supp-accepted-calibration}
\end{equation}
If calibration and test pairs are exchangeable from the intended inlier distribution, and the gate is a fixed measurable function of the image and frozen-model outputs, then accepted calibration pairs and an accepted inlier test pair are exchangeable conditional on acceptance. Consequently, when localization is disabled, ordinary split conformal prediction applied to $\mathcal{D}_{\mathrm{cal}}^{A}$ provides the usual finite-sample coverage statement for accepted inlier examples, conditional on there being at least one accepted calibration example. With localized weights, CALCoDe uses the construction in Eq.~(12) of the manuscript and inherits the assumptions stated in Section \emph{Localized Conformal Calibration}. The reported marginal, class-wise, subgroup, and bootstrap diagnostics quantify empirical behavior beyond the exchangeable inlier setting.

\subsection{Cross-Fitted Class-Tail Discovery}
\label{sec:supp-tail-discovery}

The accepted validation set is partitioned into $F$ stratified folds. Each validation example is evaluated by a pilot localized conformal rule fitted on the other $F-1$ folds. The resulting out-of-fold coverage indicators determine $\mathcal{Y}_{\mathrm{tail}}$. Hence $\mathcal{Y}_{\mathrm{tail}}$ is measurable with respect to $\mathcal{D}_{\mathrm{ref}}$, $\mathcal{D}_{\mathrm{val}}$, and $\mathcal{D}_{\mathrm{gate}}$, and is fixed before $\mathcal{D}_{\mathrm{cal}}$ is accessed. Classes with fewer than $n_{\min}$ accepted validation examples are included conservatively rather than declared reliable from insufficient evidence.

\subsection{Protected-Class Coverage and Structural Monotonicity}
\label{sec:supp-protected-coverage}

Fix a protected class $y\in\mathcal{Y}_{\mathrm{tail}}$. Let $n_y$ be the number of accepted class-$y$ examples in $\mathcal{D}_{\mathrm{cal}}$, let $r_1^{(y)},\ldots,r_{n_y}^{(y)}$ be their true-label nonconformity scores, and set
\begin{equation}
k_y=\left\lceil(n_y+1)(1-\delta_y)\right\rceil,
\qquad
q_y=r_{(k_y)}^{(y)},
\label{eq:supp-class-quantile}
\end{equation}
where $\lceil t\rceil$ denotes the smallest integer greater than or equal to $t$, and an additional value $+\infty$ is appended before taking the order statistic. Suppose the frozen gate and score are fixed, and an accepted class-$y$ test example is exchangeable with the accepted class-$y$ calibration examples. Conditional on the preceding selection stages,
\begin{equation}
\Pr\{s_{\mathrm{APS}}(X,y)\leq q_y\mid Y=y,\,A(X)=1,\,y\in\mathcal{Y}_{\mathrm{tail}}\}
\geq 1-\delta_y.
\label{eq:supp-protected-coverage}
\end{equation}
This follows from the exchangeable rank of the test score among the $n_y+1$ scores. Cross-fitted discovery is essential here: the event $y\in\mathcal{Y}_{\mathrm{tail}}$ does not reuse the final class-$y$ calibration scores that define $q_y$.

CALCoDe takes $\tau_y^{\mathrm{tail}}$ to be at least $q_y$, optionally enlarged by a class-local threshold. For any protected class $y$, the final threshold is
\begin{equation}
\tau_{\mathrm{CALCoDe}}(x,y)=\max\{\tau_{\delta,h}(x),\tau_y^{\mathrm{tail}}\},
\label{eq:supp-tail-threshold}
\end{equation}
so class membership is monotone relative to localized conformal prediction:
\begin{equation}
y\in C_{\delta,h}^{\mathrm{loc}}(x)\quad\Rightarrow\quad y\in C_{\mathrm{CALCoDe}}(x).
\label{eq:supp-tail-monotonicity}
\end{equation}
The guard therefore cannot remove a protected true class that the localized set already included, and Eq.~\eqref{eq:supp-protected-coverage} is inherited from the class-global component. The guarantee is at the prespecified guard level $1-\delta_y$, which may differ from the pooled localized target. Neither statement supplies distribution-free protection for a target cohort that violates the accepted-example exchangeability condition.

\section{Relation to Recent Approaches}
\label{sec:supp-recent-approaches}

Medical VLM adaptation usually optimizes the text--image interface while retaining a top-1 output \citep{liu2025kpl,koleilat2025biomedcoop,miao2026biodpp,shao2026vmfcoop,cui2026biomedccpl}. CALCoDe acts after that interface: it converts frozen class evidence into a selective prediction set and targets the least reliable class rather than average accuracy.

Calibration-aware prompt learning improves confidence during adaptation \citep{basu2025calibprompt}. Conf-OT, LCP-VLM, and LATA pursue transfer or local set efficiency \citep{silvarodriguez2025confot,fuchs2026lcpvlm,bozorgtabar2026lata}; $\mathrm{CP}_{r\text{-value}}$ models repeated-score instability \citep{zeng2026cpr}. CALCoDe uses prompt variability only in its robust evidence and audit layers. Its class-tail guard is driven by held-out class-wise coverage failure and calibrated after that failure is discovered.

TACP and sTACP are the closest class-tail baselines \citep{liu2026tacp}. Their penalties are derived from source-label frequency, which is appropriate when the long-tailed label distribution determines the vulnerable classes. CALCoDe defines fragility operationally through out-of-fold coverage. This distinction matters under external site shift, where the target-limiting class can be common in the source cohort or can change across frozen backbones.

CALCoDe is also distinct from Mondrian conformal prediction \citep{vovk2005algorithmic}. Mondrian calibration stratifies by class and is a natural way to seek class-conditional reliability, but it applies separate calibration rigidly across all classes, including classes that show no validation evidence of tail risk. Under clinical shift this can inflate set sizes broadly or still fail when the limiting issue is local support rather than class frequency alone. CALCoDe uses a one-sided adaptive guard: localized calibration remains the default, and additional class-conditional slack is introduced only for validation-detected tail-risk classes. The guard therefore targets weak-class undercoverage without converting every class into an independent worst-case calibration problem.

Relative to standard OOD detection and simple distance gates, CALCoDe couples rejection with calibrated set construction. OOD scores usually produce a binary accept--reject decision or anomaly ranking, and recent medical OOD benchmarks show that natural-image conclusions transfer poorly to medical imaging \citep{gutbrod2025openmibood}. CALCoDe uses a validation-audited support score for deferral and embedding-localized conformal calibration for accepted cases. Its assumptions remain tied to the stated exchangeability conditions. Empirically, the method combines robust prompt-derived APS nonconformity, local calibration, validation-discovered class-tail protection, and calibrated abstention for medical VLM classification.

\section{Experimental Setup}
\label{sec:supp-experimental-setup}

\subsection{Implementation and Data Splits}

The primary external-shift dermatology experiments evaluate four frozen image-text backbones: Microsoft BiomedCLIP, OpenAI CLIP ViT-B/32, PubMedCLIP ViT-B/32, and Google MedSigLIP-448. The exact model identifiers, preprocessing choices, and resolved configuration files are saved with each run. No backbone is fine-tuned. Each dermatology class is represented using five prompt templates:
\[
\begin{gathered}
\text{``a dermoscopic image showing \{label\}.''}\\
\text{``a clinical dermatology image consistent with \{label\}.''}\\
\text{``a skin lesion diagnosed as \{label\}.''}\\
\text{``a medical image of \{label\}.''}\\
\text{``a pigmented lesion image showing \{label\}.''}
\end{gathered}
\]
Experiments use two compute environments. The Windows 10 workstation contains two Intel Xeon Gold 6226R processors (32 physical cores in total) and two NVIDIA RTX A4500 GPUs with 20\,GB memory each; it uses NVIDIA driver 553.35, CUDA 12.4, Python 3.10.20, and PyTorch 2.5.1 with CUDA 12.1 support. The Linux DGX A100 node contains eight NVIDIA A100-SXM4 GPUs with 40\,GB memory each; each Slurm job reserves one GPU and uses NVIDIA driver 535.54.03, CUDA 12.2, Python 3.10.12, and PyTorch 2.5.1 with CUDA 12.1 support. Feature extraction uses deterministic evaluation mode. The mini-batch size is 32 for ViT-B/32-scale backbones on the Windows workstation and 16 for MedSigLIP-448 because it uses 448-resolution inputs. Camelyon17 feature extraction for BiomedCLIP and OpenAI CLIP uses one A100 GPU and batch size 128. Batch size changes throughput only because all backbones remain frozen and evaluation is deterministic. The reported CALCoDe runs use APS nonconformity from prompt probabilities trimmed with $\kappa=1$, $k=10$ neighbors for the global audit distance, an automatically selected localization bandwidth, local $\eta$ search grid size 1001, five stratified folds for validation-only class-tail discovery, discovery threshold $0.90$, and $n_{\min}=10$. Classes below $n_{\min}$ are protected by default. The final class-tail guard level is $0.95$, fixed before test evaluation. Unless otherwise stated, results use pooled target coverage $0.95$ and deferral level $\alpha_{\mathrm{def}}=0.05$. The utility-policy diagnostic searches target coverages $\{0.90,0.925,0.95\}$ and deferral levels $\{0.01,0.02,0.05,0.10\}$ on the validation split, fixes the selected operating point, and then reports held-out test metrics from the cached predictions.

\paragraph{HAM10000 in-distribution evaluation.}
The 10,015-image HAM10000 collection, released as the training data for ISIC 2018 Task 3 (lesion diagnosis), is split by lesion group into reference, validation, gate, conformal-calibration, and test roles. The resulting split sizes are: reference 2040, validation 1935, gate 1986, conformal calibration 2050, and test 2004.

\paragraph{ISIC 2019 external-shift evaluation.}
For the external-shift setting, HAM10000 provides the reference, validation, gate, and conformal-calibration data. ISIC 2019 test images form the evaluation cohort. The combined experiment contains 16249 images, with split sizes: reference 2040, validation 1935, gate 1986, conformal calibration 2050, and test 8238. Among the ISIC 2019 test images, 6026 are inlier examples mapped to the seven HAM10000 labels, while 2212 are OOD examples from classes outside the HAM10000 label space.

\paragraph{PAD-UFES-20 external-shift evaluation.}
For the second external-shift setting, HAM10000 supplies the reference, validation, gate, and conformal-calibration data, while PAD-UFES-20 supplies the evaluation cohort. The processed PAD-UFES-20 evaluation contains 2106 images mapped to the dermatology label set used in the conformal experiments. This cohort serves as an external-domain stress test; OOD detection lies outside the associated claim.

\subsection{Backbone Geometry Audit}
The PAD-UFES-20 shift reveals strong coupling between the audited gate and the frozen embedding geometry. An audit from cached features quantifies that coupling. Let $d_k(x,\mathcal{R})$ be the median cosine distance from $x$ to its $k=10$ nearest neighbors in the HAM10000 reference set $\mathcal{R}$. The global shift ratio is the median of $d_k$ on PAD-UFES-20 divided by the corresponding median on held-out HAM10000 source examples. The class-local shift ratio uses only reference neighbors with the same mapped class label. The audit also reports a linear source-versus-target probe balanced accuracy, effective rank of the HAM10000 source embedding covariance, and the top principal-component variance share. All diagnostics are computed after feature normalization and play no role in method training or selection.

\begin{table}[H]
\centering
\scriptsize
\setlength{\tabcolsep}{4pt}
\begin{tabular}{lcccccccc}
\hline
Backbone & Global shift & Class shift & Probe bal. acc. & Eff. rank & Top1 var. & Median $p_{\mathrm{audit}}$ & Def. & Set \\
\hline
BiomedCLIP & 5.518 & 5.800 & 0.991 & 38.529 & 0.132 & 0.004 & 0.832 & 4.927 \\
OpenAI CLIP ViT-B/32 & 2.420 & 2.640 & 0.998 & 47.994 & 0.198 & 0.388 & 0.009 & 4.991 \\
PubMedCLIP ViT-B/32 & 2.134 & 2.349 & 0.997 & 39.923 & 0.199 & 0.082 & 0.403 & 4.923 \\
MedSigLIP-448 & 2.804 & 3.027 & 1.000 & 47.123 & 0.169 & 0.778 & 0.042 & 4.986 \\
\hline
\end{tabular}
\caption{Detailed PAD-UFES-20 embedding-geometry audit. BiomedCLIP places PAD-UFES-20 farthest from the HAM10000 reference manifold and has the smallest audited p-values. OpenAI CLIP and MedSigLIP defer less, with accepted set sizes remaining nearly full; low deferral is therefore insufficient evidence of diagnostic specificity.}
\label{tab:supp-pad-geometry}
\end{table}

At the fixed comparison operating point, the median audited inlier-support p-value on PAD-UFES-20 is 0.004 for BiomedCLIP, 0.388 for OpenAI CLIP ViT-B/32, 0.082 for PubMedCLIP, and 0.778 for MedSigLIP-448. OpenAI CLIP and MedSigLIP defer less often, yet their accepted prediction sets remain close to the full five-class label space and their accepted singleton rate is zero. Deferral, set size, and singleton rate therefore require joint interpretation under this shift.

\paragraph{Camelyon17-WILDS stress test.}
Camelyon17-WILDS is used as a supplementary hospital-shift stress test in histopathology. The source hospitals provide reference, validation, gate, and calibration splits, while hospital 2 is held out as the target test domain. The binary label space contains non-tumor lymph node tissue and metastatic tumor tissue. To keep the experiment computationally controlled, the held-out test set is balanced to 5000 examples per class.

\subsection{Metrics, Confidence Intervals, and Audit Stability}

\paragraph{Metrics.}
Coverage is the fraction of accepted inlier examples for which the true class appears in the prediction set. Set size is the average number of labels returned on accepted examples. Deferral is the fraction of examples rejected by the deferral gate. In-def. and OOD-def. are deferral rates for inlier and OOD examples, respectively. Selective accuracy is top-1 accuracy on accepted inlier examples. ECE is expected calibration error on accepted examples. Worst-class coverage is the minimum accepted coverage across inlier disease classes.

\begin{figure}[H]
\centering
\begin{subfigure}[t]{0.485\linewidth}
  \centering
  \includegraphics[width=\linewidth]{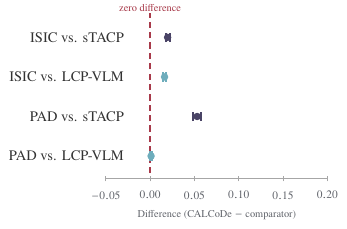}
  \caption{Accepted-coverage difference.}
  \label{fig:supp-paired-coverage}
\end{subfigure}\hfill
\begin{subfigure}[t]{0.485\linewidth}
  \centering
  \includegraphics[width=\linewidth]{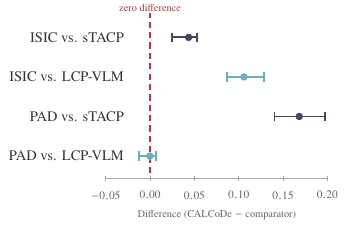}
  \caption{Worst-class accepted-coverage difference.}
  \label{fig:supp-paired-worst-class}
\end{subfigure}
\caption{Paired CALCoDe-minus-comparator differences after predictions are fixed, macro-averaged over four backbones. Bars show 95\% class-stratified bootstrap intervals. CALCoDe improves both metrics over sTACP on both shifts and over LCP-VLM on ISIC 2019; differences from LCP-VLM on PAD-UFES-20 include zero.}
\label{fig:supp-paired-reliability}
\end{figure}

\paragraph{Bootstrap confidence intervals.}
Per-setting confidence intervals use 500 bootstrap resamples over the saved per-example prediction files. The recent-baseline comparisons use 1000 paired resamples, stratified by true class within each backbone; each replicate recomputes the metric difference and then averages it over the four backbones. Both procedures are applied after all model outputs, gates, and prediction sets have been fixed. They measure sampling variation in the empirical metrics without changing any model decision. Figure~\ref{fig:supp-paired-reliability} visualizes the paired differences, and Table~\ref{tab:supp-paired-bootstrap} provides their numerical intervals. The released diagnostic package includes:
\begin{flushleft}
\footnotesize\ttfamily
recent\_baselines\_summary.csv\\
recent\_baselines\_dataset\_means.csv\\
recent\_baselines\_classwise.csv\\
recent\_baselines\_deltas.csv\\
recent\_baselines\_paired\_bootstrap.csv\\
calcode\_matched\_ablation\_summary.csv\\
calcode\_matched\_ablation\_means.csv\\
calcode\_matched\_ablation\_deltas.csv\\
calcode\_matched\_ablation\_reference\_check.csv\\
audit\_score\_stability.csv\\
geometry\_summary.csv\\
distance\_quantiles.csv\\
pca\_domain\_projection.csv\\
gamma\_sensitivity\_average.csv\\
utility\_calcode\_comparison.csv\\
utility\_calcode\_dataset\_means.csv\\
validation\_selected\_risk\_delta\_summary.csv\\
conservative\_xmodal\_budget\_summary.csv\\
camelyon\_binary\_informative\_utility\_cov0.95\_def0.05.csv\\
ann\_knn\_scaling.csv
\end{flushleft}

\begin{table}[H]
\centering
\small
\setlength{\tabcolsep}{5pt}
\begin{tabular}{llcc}
\hline
Shift & Comparator & $\Delta$ accepted cov. & $\Delta$ worst-class cov. \\
\hline
ISIC 2019 & sTACP & $0.0198\ [0.0175,0.0222]$ & $0.0435\ [0.0248,0.0529]$ \\
ISIC 2019 & LCP-VLM & $0.0162\ [0.0147,0.0178]$ & $0.1059\ [0.0870,0.1285]$ \\
PAD-UFES-20 & sTACP & $0.0531\ [0.0487,0.0573]$ & $0.1683\ [0.1403,0.1979]$ \\
PAD-UFES-20 & LCP-VLM & $0.0011\ [-0.0004,0.0023]$ & $-0.0003\ [-0.0126,0.0071]$ \\
\hline
\end{tabular}
\caption{Paired macro-average differences, CALCoDe minus comparator, with 95\% class-stratified bootstrap intervals. Positive values favor CALCoDe.}
\label{tab:supp-paired-bootstrap}
\end{table}

\paragraph{Audit-score stability.}
Since the gate diagnostic is selected on validation data, we also bootstrap the validation split and re-run the score-selection step for the core candidate diagnostics: distance, energy, maximum-softmax confidence, prompt disagreement, and their fused-rank score. Table~\ref{tab:supp-audit-stability} reports the diagnostic selected by the full validation split, its validation AUROC for top-1 failure detection, the AUROC margin over the second-best diagnostic, and the fraction of 1000 bootstrap resamples that select the same diagnostic. Six of the eight settings are stable above 0.95. The two unstable cases are both OpenAI CLIP settings, where all candidate diagnostics have weak failure-detection signal and the AUROC margin is below 0.01. The audit is therefore interpreted as a data-dependent gate selector, with no universal score-ranking theorem implied.

\begin{table}[H]
\centering
\scriptsize
\setlength{\tabcolsep}{5pt}
\begin{tabular}{llcccc}
\hline
Dataset & Backbone & Selected score & Val. AUROC & AUROC margin & Bootstrap freq. \\
\hline
ISIC 2019 & BiomedCLIP & energy & 0.681 & 0.033 & 0.973 \\
ISIC 2019 & OpenAI CLIP ViT-B/32 & MSP & 0.570 & 0.003 & 0.589 \\
ISIC 2019 & PubMedCLIP ViT-B/32 & prompt & 0.730 & 0.026 & 0.986 \\
ISIC 2019 & MedSigLIP-448 & energy & 0.578 & 0.123 & 1.000 \\
PAD-UFES-20 & BiomedCLIP & fused rank & 0.699 & 0.019 & 0.956 \\
PAD-UFES-20 & OpenAI CLIP ViT-B/32 & MSP & 0.473 & 0.009 & 0.627 \\
PAD-UFES-20 & PubMedCLIP ViT-B/32 & prompt & 0.757 & 0.056 & 1.000 \\
PAD-UFES-20 & MedSigLIP-448 & energy & 0.581 & 0.086 & 0.994 \\
\hline
\end{tabular}
\caption{Audit-score stability over 1000 validation bootstrap resamples. MSP denotes negative maximum softmax probability, so larger values indicate weaker confidence. Bootstrap frequency is the probability that the resampled validation split selects the same diagnostic as the full validation split.}
\label{tab:supp-audit-stability}
\end{table}

\subsection{Component Ablations and Sensitivity}

\paragraph{Ablation summary.}
Table~\ref{tab:supp-ablation-average} reports the component ablation associated with Section \emph{Q3: Which Components Drive Class-Tail Reliability?} of the manuscript. The values are averaged over HAM10000$\rightarrow$ISIC 2019 and HAM10000$\rightarrow$PAD-UFES-20 across the four frozen backbones.

\begin{table}[H]
\centering
\scriptsize
\setlength{\tabcolsep}{2pt}
\begin{tabular}{lcccccc}
\hline
Variant & Cov. & Worst cov. & Set & Def. & Marg. target & Worst target \\
\hline
Without class-tail guard & 0.985 & 0.917 & 5.68 & 0.227 & 8/8 & 5/8 \\
Without adaptive discovery & 0.998 & 0.997 & 5.98 & 0.226 & 8/8 & 8/8 \\
Without support audit & 0.995 & 0.983 & 5.87 & 0.000 & 8/8 & 8/8 \\
Without localization & 0.976 & 0.900 & 5.77 & 0.226 & 7/8 & 3/8 \\
Without prompt trimming & 0.992 & 0.970 & 5.85 & 0.226 & 8/8 & 6/8 \\
\hline
\rowcolor{CalCoDE-Row}\textbf{CALCoDe} & 0.994 & 0.979 & 5.85 & 0.226 & 8/8 & 8/8 \\
\hline
\end{tabular}
\caption{Matched CALCoDe ablations averaged over the two dermatology external-shift datasets and four backbones. Each variant changes one component of the verified APS implementation. Shading identifies the complete method.}
\label{tab:supp-ablation-average}
\end{table}

Removing the class-tail guard or localization lowers worst-class coverage to 0.917 and 0.900, respectively. Prompt trimming has a smaller average effect but prevents additional setting-level failures: the untrimmed variant reaches the 0.95 worst-class target in six of eight settings, compared with eight of eight for CALCoDe. Guarding all classes reaches 0.997 worst-class coverage while adding 0.13 labels to the mean set, which identifies adaptive discovery as an efficiency mechanism. The no-audit variant retains similar accepted-example coverage and sets deferral to zero; the audit governs escalation rather than generating the class-tail coverage gain. The released matched-ablation summary also records each setting and verifies that the complete-method rows reproduce the primary results exactly.

\paragraph{Tail-risk margin sensitivity.}
CALCoDe marks a class as tail-risk when its validation coverage falls below $1-\delta-\gamma$. Table~\ref{tab:supp-gamma-sensitivity} varies $\gamma$ with the class-tail guard held at 0.90, keeping the frozen backbones, calibration splits, audited gate, and pooled target fixed. The selected value $\gamma=0.05$ corresponds to a validation discovery boundary of $0.90$ at pooled target coverage $0.95$. The broad plateau establishes that the discovered class set is not sensitive to small changes in the margin. A separate confirmatory run in Table~\ref{tab:supp-guard-level} sets the final guard to 0.95, as used by the reported CALCoDe method.

\begin{table}[H]
\centering
\small
\setlength{\tabcolsep}{5pt}
\begin{tabular}{lcccc}
\hline
$\gamma$ & Cov. & Worst cov. & Set & Def. \\
\hline
0.00 & 0.993 & 0.978 & 5.850 & 0.226 \\
0.02 & 0.993 & 0.978 & 5.850 & 0.226 \\
0.04 & 0.993 & 0.978 & 5.838 & 0.226 \\
\textbf{0.05} & 0.993 & 0.978 & 5.838 & 0.226 \\
0.06 & 0.993 & 0.978 & 5.838 & 0.226 \\
0.08 & 0.992 & 0.966 & 5.814 & 0.226 \\
0.10 & 0.992 & 0.966 & 5.813 & 0.226 \\
\hline
\end{tabular}
\caption{Sensitivity to the validation tail-risk margin $\gamma$ at class-tail guard level 0.90, averaged over the eight dermatology settings. Larger $\gamma$ marks fewer validation classes as tail-risk classes; bold marks the fixed discovery margin.}
\label{tab:supp-gamma-sensitivity}
\end{table}

\paragraph{Guard-level confirmation.}
The final method uses a 0.95 class-tail guard, matching the pooled target. Table~\ref{tab:supp-guard-level} compares this choice with the earlier 0.90 guard while holding the validation-discovered class sets, audit, localization, and all data splits fixed. Raising the guard improves ISIC 2019 worst-class coverage by 0.0035 and increases its average set size by 0.028 labels. PAD-UFES-20 coverage and deferral are unchanged to the displayed precision.

\begin{table}[H]
\centering
\footnotesize
\setlength{\tabcolsep}{3.5pt}
\begin{tabular}{lccccccc}
\hline
Guard & \multicolumn{3}{c}{ISIC 2019} & \multicolumn{3}{c}{PAD-UFES-20} & Mean def. \\
\cline{2-4}\cline{5-7}
level & Cov. & Worst cov. & Set & Cov. & Worst cov. & Set & \\
\hline
0.90 & 0.988 & 0.966 & 6.684 & 0.999 & 0.989 & 4.991 & 0.226 \\
\rowcolor{CalCoDE-Row}\textbf{0.95} & \textbf{0.989} & \textbf{0.970} & 6.712 & \textbf{0.999} & \textbf{0.989} & 4.991 & 0.226 \\
\hline
\end{tabular}
\caption{Confirmatory comparison of class-tail guard levels, macro-averaged over four backbones. The reported experiments use the 0.95 row.}
\label{tab:supp-guard-level}
\end{table}

\subsection{Recent Baseline Implementation}

All recent baselines are evaluated from the same cached prompt-level probabilities, frozen embeddings, class names, and split assignments used by CALCoDe. The matched-prompt $\mathrm{CP}_{r\text{-value}}$ baseline follows the released nonparametric greedy majority ordering: the five prompt variants form repeated score realizations for both calibration true labels and test candidates. TACP and sTACP use HAM10000 source-class priors and APS scores. Their regularization weight and rank parameter are selected by five-fold source validation, subject to a 0.01 marginal-coverage tolerance, before the disjoint calibration split supplies the final threshold. Neither method accesses target labels.

The LATA comparison applies its Laplacian graph-refinement component to the pooled source and unlabeled target embeddings, using $k=15$, graph weight 0.35, and eight refinement iterations. The separately trained ViLU head is unavailable in this frozen-backbone setting, so every table names the method ``LATA graph refinement (no ViLU).'' These results characterize the graph component and do not support a claim about the complete LATA system. LCP-VLM, Conf-OT, Mondrian CP, local CP, and APS use the same calibration split and target level as CALCoDe. Empty sets are counted as non-informative outputs; methods without an explicit rejection mechanism show ``--'' for deferral in Table~1 of the manuscript.

\section{Camelyon17-WILDS Hospital-Shift Stress Test}
\label{sec:supp-camelyon17}

Table~\ref{tab:supp-camelyon17} reports the Camelyon17-WILDS hospital-shift stress test, which is separate from the primary dermatology benchmarks and evaluates CALCoDe without class-tail guarding. For weaker zero-shot backbones, the prediction sets approach the full binary label set, yielding conservative behavior with limited diagnostic specificity. BiomedCLIP achieves the most informative Camelyon17 tradeoff, with accepted coverage 0.981, average set size 1.746, and worst-class coverage 0.962.

\begin{table}[H]
\centering
\scriptsize
\setlength{\tabcolsep}{4pt}
\begin{tabular}{lcccccc}
\hline
Method & Cov.$\uparrow$ & Set$\downarrow$ & Def.$\downarrow$ & Sel. acc.$\uparrow$ & ECE$\downarrow$ & Worst cov.$\uparrow$ \\
\backbonerow{BiomedCLIP}
Split CP & 0.945 & 1.720 & 0.000 & 0.621 & 0.154 & 0.902 \\
Local CP & 0.971 & 1.816 & 0.031 & 0.610 & 0.163 & 0.944 \\
Conf-OT & 1.000 & 2.000 & 0.000 & 0.691 & 0.094 & 1.000 \\
CALCoDe base (no tail guard) & 0.981 & 1.746 & 0.177 & 0.656 & 0.120 & 0.962 \\
\backbonerow{OpenAI CLIP ViT-B/32}
Split CP & 0.892 & 1.799 & 0.000 & 0.426 & 0.120 & 0.810 \\
Local CP & 0.992 & 1.985 & 0.000 & 0.426 & 0.120 & 0.983 \\
Conf-OT & 1.000 & 2.000 & 0.000 & 0.494 & 0.141 & 1.000 \\
CALCoDe base (no tail guard) & 0.986 & 1.970 & 0.026 & 0.425 & 0.122 & 0.973 \\
\backbonerow{PubMedCLIP ViT-B/32}
Split CP & 0.812 & 1.572 & 0.000 & 0.502 & 0.126 & 0.625 \\
Local CP & 0.984 & 1.958 & 0.000 & 0.502 & 0.126 & 0.969 \\
Conf-OT & 1.000 & 2.000 & 0.000 & 0.503 & 0.134 & 1.000 \\
CALCoDe base (no tail guard) & 0.987 & 1.954 & 0.069 & 0.506 & 0.130 & 0.974 \\
\backbonerow{MedSigLIP-448}
Split CP & 0.955 & 1.953 & 0.000 & 0.404 & 0.267 & 0.910 \\
Local CP & 0.999 & 1.999 & 0.000 & 0.404 & 0.267 & 0.999 \\
Conf-OT & 1.000 & 2.000 & 0.000 & 0.337 & 0.341 & 1.000 \\
CALCoDe base (no tail guard) & 0.998 & 1.998 & 0.025 & 0.404 & 0.267 & 0.996 \\
\hline
\end{tabular}
\caption{Camelyon17-WILDS hospital-shift stress test. Metrics are computed on accepted examples from held-out hospital 2. In the binary setting, set size near 2 indicates conservative full-set behavior.}
\label{tab:supp-camelyon17}
\end{table}

\section{Utility and Deployment Diagnostics}
\label{sec:supp-utility}

\paragraph{Reliability--utility profile.}
Table~\ref{tab:supp-recent-utility} complements Table~1 of the manuscript with two cohort-level utility measures. Accepted full-set rate is the fraction of accepted examples assigned every available label. Autonomous informative rate is the fraction of the full cohort that is accepted with a set smaller than the complete label space. The target-hit column counts the eight dataset--backbone settings with observed accepted coverage and worst-class coverage at least 0.95. CALCoDe has the most consistent class-tail reliability, while its autonomous informative rate is low on PAD-UFES-20. Conf-OT and $\mathrm{CP}_{r\text{-value}}$ return more informative sets but miss the reliability target in most settings. LCP-VLM and CALCoDe both approach complete five-class sets on PAD-UFES-20.

\begin{table}[H]
\centering
\scriptsize
\setlength{\tabcolsep}{4pt}
\begin{tabular}{lccccc}
\hline
& \multicolumn{2}{c}{ISIC 2019} & \multicolumn{2}{c}{PAD-UFES-20} & Target hits \\
\cline{2-3}\cline{4-5}
Method & Full set & Auto. info. & Full set & Auto. info. & Marg./worst \\
\hline
APS & 0.265 & 0.735 & 0.504 & 0.496 & 3/1 \\
Mondrian CP & 0.425 & 0.575 & 0.362 & 0.638 & 2/0 \\
Conf-OT APS & 0.000 & 1.000 & 0.000 & 1.000 & 0/0 \\
$\mathrm{CP}_{r\text{-value}}$ & 0.375 & 0.625 & 0.416 & 0.584 & 2/0 \\
TACP-APS & 0.656 & 0.344 & 0.700 & 0.300 & 6/2 \\
sTACP-APS & 0.796 & 0.204 & 0.618 & 0.382 & 7/1 \\
LATA graph refinement (no ViLU) & 0.325 & 0.672 & 0.534 & 0.466 & 4/2 \\
LCP-VLM APS & 0.714 & 0.286 & 0.992 & 0.008 & 8/5 \\
\rowcolor{CalCoDE-Row}\textbf{CALCoDe} & 0.790 & 0.184 & 0.993 & 0.003 & \textbf{8/8} \\
\hline
\end{tabular}
\caption{Reliability and utility macro-averages over four backbones. Full set is the accepted full-label-set rate; Auto. info. is the full-cohort autonomous informative rate. Target hits count settings with empirical marginal/worst-class accepted coverage at least 0.95. Higher target hits and Auto. info. are preferable; lower Full set is preferable.}
\label{tab:supp-recent-utility}
\end{table}

\paragraph{Cost-sensitive risk.}
Eq.~(2) of the manuscript defines selective decision risk. Its clinical costs depend on the deployment setting. For the empirical summary in Table~\ref{tab:supp-risk}, let $\hat d$ denote the deferral rate, $\widehat C$ accepted coverage, $\widehat C_{\mathrm{w}}$ worst-class accepted coverage, $\widehat S$ accepted set size, and $K$ the number of classes. We evaluate
\begin{equation}
\widehat R=(1-\hat d)\left[c_{\mathrm{err}}(1-\widehat C)+\frac{c_{\mathrm{err}}}{2}(1-\widehat C_{\mathrm{w}})+c_{\mathrm{set}}\frac{\widehat S-1}{K-1}\right]+c_{\mathrm{def}}\hat d .
\label{eq:supp-cost-risk}
\end{equation}
The values are averaged over the eight dermatology external-shift settings at the fixed operating point reported in Section \emph{Evaluation Setup} of the manuscript. The resulting risks identify the operating regimes suited to each method. CALCoDe is favored when weak-class protection has substantial cost; LCP-VLM is favored when deferral is assigned a very high cost.

\begin{table}[H]
\centering
\scriptsize
\setlength{\tabcolsep}{4pt}
\begin{tabular}{llcccc}
\hline
Scenario & Cost vector $(c_{\mathrm{err}},c_{\mathrm{set}},c_{\mathrm{def}})$ & Conf-OT & LCP-VLM & No class-tail guard & CALCoDe \\
\hline
Balanced & $(10,1,1)$ & 5.479 & 1.464 & 1.516 & \best{1.101} \\
Safety-heavy & $(25,1,0.5)$ & 12.738 & 2.230 & 2.263 & \best{1.170} \\
Deferral-expensive & $(10,1,5)$ & 5.479 & \best{1.464} & 2.422 & \second{2.007} \\
Efficiency-heavy & $(10,3,1)$ & 6.757 & 3.369 & 2.948 & \best{2.608} \\
\hline
\end{tabular}
\caption{Cost-sensitive risk averaged over HAM10000$\rightarrow$ISIC 2019 and HAM10000$\rightarrow$PAD-UFES-20 across four backbones. Lower is better. CALCoDe is strongest in balanced, safety-heavy, and efficiency-heavy regimes; LCP-VLM is preferred when any deferral is much more costly than class-tail undercoverage.}
\label{tab:supp-risk}
\end{table}

\paragraph{Validation-selected utility policy.}
The fixed comparison operating point supports method comparison, while a deployment site may prefer a different point on the calibrated grid after specifying its clinical costs. Table~\ref{tab:supp-risk-frontier} reports the average held-out change obtained by selecting the audited base-layer operating point on the validation split and then evaluating the selected policy on the external-shift test cohort. The selector minimizes the risk above over the saved grid, subject to an accepted-coverage floor of 0.95, a minimum of 50 accepted validation examples when feasible, and a minimum informative-decision rate of 0.02. The feasibility constraint is relaxed when no grid point satisfies it, which occurs in three of eight dataset--backbone settings because the validation split itself exposes limited utility. The deployment policy chooses among calibrated operating points after clinical costs are specified and adds no statistical guarantee.

\begin{table}[H]
\centering
\scriptsize
\setlength{\tabcolsep}{5pt}
\begin{tabular}{lcccccc}
\hline
Scenario & $\Delta\,\mathrm{risk}\downarrow$ & $\Delta\,\mathrm{cov.}\uparrow$ & $\Delta\,\mathrm{worst}\uparrow$ & $\Delta\,\mathrm{set}\downarrow$ & $\Delta\,\mathrm{def.}\downarrow$ & Chosen def. \\
\hline
Balanced & -0.037 & -0.001 & -0.014 & -0.044 & +0.039 & 0.364 \\
Safety-heavy & -0.138 & -0.001 & -0.013 & -0.044 & +0.059 & 0.385 \\
Deferral-expensive & -0.558 & +0.001 & +0.012 & +0.044 & -0.150 & 0.175 \\
Efficiency-heavy & -0.154 & -0.001 & -0.013 & -0.044 & +0.059 & 0.385 \\
\hline
\end{tabular}
\caption{Validation-selected utility policy for the audited base layer, averaged over the eight dermatology external-shift settings. Deltas are held-out test differences relative to the fixed comparison point $\alpha_{\mathrm{def}}=0.05$. Negative risk and deferral deltas are improvements; positive coverage and worst-class-coverage deltas are improvements.}
\label{tab:supp-risk-frontier}
\end{table}

\paragraph{Cross-modal disagreement diagnostic.}
We also evaluated cross-modal disagreement as a candidate audit score. The score compares a confident text-induced class prediction with weak class-local image support. A conservative selector allowed this score, or a fused rank containing it, to replace the baseline audit only when it improved validation failure-detection AUROC by at least 0.02 and did not exceed a validation deferral budget. Table~\ref{tab:supp-xmodal-negative} shows the target-domain switches that passed this validation filter. Two switches are benign or helpful, while two substantially increase target deferral. The observed target-domain instability supports reporting cross-modal disagreement as a diagnostic analysis. CALCoDe uses the remaining audited reliability candidates in its decision rule.

\begin{table}[H]
\centering
\scriptsize
\setlength{\tabcolsep}{4pt}
\begin{tabular}{llrrrr}
\hline
Dataset/backbone & Selected audit & $\Delta\,\mathrm{worst}$ & $\Delta\,\mathrm{set}$ & $\Delta\,\mathrm{def.}$ & $\Delta\,\mathrm{sel.\ acc.}$ \\
\hline
ISIC 2019 / MedSigLIP-448 & disagreement & -0.018 & -0.103 & -0.003 & +0.004 \\
PAD-UFES-20 / BiomedCLIP & fused disagreement & +0.007 & +0.020 & -0.036 & +0.003 \\
PAD-UFES-20 / OpenAI CLIP & disagreement & +0.004 & +0.002 & +0.169 & +0.013 \\
PAD-UFES-20 / MedSigLIP-448 & disagreement & -0.008 & -0.024 & +0.547 & -0.180 \\
\hline
\end{tabular}
\caption{Accepted switches under conservative cross-modal disagreement selection. Deltas are relative to the baseline audited gate at the same operating point. The validation-budget rule permits harmful target-domain deferral increases, so the disagreement analysis remains exploratory and outside the proposed method.}
\label{tab:supp-xmodal-negative}
\end{table}

\paragraph{Binary full-set utility.}
In a binary task, returning the full set $\mathcal{Y}$ is mathematically covered yet diagnostically uninformative. Table~\ref{tab:supp-binary-utility} reports both the original binary utility diagnostic and the stricter deployment rule in Eq.~(19) of the manuscript, which converts empty or full binary sets into deferrals. The stress test exposes a real limitation: except for BiomedCLIP, most frozen zero-shot backbones produce few informative binary decisions under the reliability constraint. In those cases, the trust layer identifies the backbone as unsuitable for autonomous hospital-shift deployment.

\begin{table}[H]
\centering
\scriptsize
\setlength{\tabcolsep}{5pt}
\begin{tabular}{lccccc}
\hline
Backbone & Cov. & Set & Orig. informative & Utility def. & Singleton cov. \\
\hline
BiomedCLIP & 0.981 & 1.746 & \best{0.209} & 0.786 & \best{0.920} \\
OpenAI CLIP ViT-B/32 & 0.986 & 1.970 & 0.030 & 0.981 & 0.612 \\
PubMedCLIP ViT-B/32 & 0.987 & 1.954 & 0.043 & 0.957 & 0.724 \\
MedSigLIP-448 & 0.998 & 1.998 & 0.002 & 0.997 & 0.115 \\
\hline
\end{tabular}
\caption{Camelyon17 binary utility diagnostic for the audited base layer. Orig. informative is the all-example singleton rate before converting full binary sets into deferrals. Utility def. is the final deferral rate after empty or full binary sets are escalated. Singleton cov. is coverage among the remaining autonomous singleton decisions.}
\label{tab:supp-binary-utility}
\end{table}

\paragraph{Compute and storage overhead.}
CALCoDe stores frozen image embeddings for the reference, calibration, and test splits. In the completed runs, the saved feature caches range from 9.4 MB to 41.2 MB per dataset-backbone pair. The exact reference sets contain 1992--2040 dermatology images and 1937 Camelyon17 images; embedding dimensions are 512 for ViT-B/32-scale CLIP/BiomedCLIP/PubMedCLIP models and 1152 for MedSigLIP-448. The global audit distance uses exact $k$-nearest-neighbor ($k$-NN) search with $k=10$ at this scale, and localized calibration uses exact cosine distances to accepted calibration embeddings. Table~\ref{tab:supp-ann-scaling} evaluates a Hierarchical Navigable Small World (HNSW) approximate-nearest-neighbor index on cached dermatology embeddings. The approximate index yields lower per-image neighbor-query latency and near-identical neighbor support distances. The HNSW experiment measures support-index scalability; all reported coverage results use exact neighbor calculations.

\begin{table}[H]
\centering
\scriptsize
\setlength{\tabcolsep}{3pt}
\begin{tabular}{llcccc}
\hline
Dataset & Backbone & Exact ms & HNSW ms & Recall@10 & P95 err. \\
\hline
ISIC 2019 & BiomedCLIP & 0.030 & 0.0067 & 0.999 & $4.47{\times}10^{-7}$ \\
ISIC 2019 & OpenAI & 0.028 & 0.0064 & 1.000 & $5.66{\times}10^{-7}$ \\
PAD-UFES & BiomedCLIP & 0.038 & 0.0098 & 0.998 & $3.87{\times}10^{-7}$ \\
PAD-UFES & MedSigLIP & 0.038 & 0.0125 & 1.000 & $5.07{\times}10^{-7}$ \\
\hline
\end{tabular}
\caption{Approximate-nearest-neighbor support-index diagnostic using HNSW with $k=10$, $M=32$, and ef-search 200. P95 err. is the 95th percentile absolute error in the median cosine-distance support statistic relative to exact $k$-NN.}
\label{tab:supp-ann-scaling}
\end{table}

\section{Supporting BiomedCLIP Results}
\label{sec:supp-biomedclip}

\begin{table}[H]
\centering
\footnotesize
\setlength{\tabcolsep}{3.5pt}
\begin{tabular}{lcccccc}
\hline
Method & Cov. & Set size & Def. & Sel. acc. & ECE & Worst-class cov. \\
\hline
Top-1 & 0.565 & 1.00 & 0.000 & 0.565 & 0.164 & 0.000 \\
Temperature scaling & 0.565 & 1.00 & 0.000 & 0.565 & \best{0.099} & 0.000 \\
Split conformal & 0.960 & 5.05 & 0.000 & 0.566 & 0.160 & 0.600 \\
APS & 0.957 & 5.20 & 0.000 & 0.566 & 0.160 & 0.400 \\
RAPS & 0.953 & 5.19 & 0.000 & 0.566 & 0.160 & 0.369 \\
Mondrian conformal & 0.956 & 6.16 & 0.000 & 0.566 & 0.160 & \best{0.930} \\
Local conformal & 0.955 & 4.20 & 0.000 & 0.566 & 0.160 & 0.615 \\
Global calibration only & 0.946 & 4.87 & 0.036 & \best{0.572} & \second{0.155} & \second{0.688} \\
Localized calibration & 0.959 & 4.28 & 0.036 & \best{0.572} & \second{0.155} & \second{0.688} \\
Localized calibration, mean support & \best{0.961} & 4.30 & \best{0.035} & \best{0.572} & \second{0.155} & \second{0.688} \\
Localized calibration, no semantic term & \second{0.960} & \best{4.08} & 0.036 & \best{0.572} & \second{0.155} & \second{0.688} \\
\hline
\end{tabular}
\caption{HAM10000 in-distribution results using frozen BiomedCLIP. Coverage, set size, selective accuracy, ECE, and worst-class coverage are computed on accepted examples.}
\label{tab:supp-ham10000}
\end{table}

\begin{table}[H]
\centering
\footnotesize
\setlength{\tabcolsep}{3.5pt}
\begin{tabular}{lccccccc}
\hline
Method & AKIEC & BCC & BKL & DF & NV & MEL & VASC \\
\hline
Split conformal & 0.211 & 0.944 & 0.824 & 0.978 & \second{0.990} & \best{0.985} & 0.760 \\
Local conformal & \second{0.741} & \best{0.965} & 0.861 & \best{0.989} & \best{0.991} & \second{0.981} & \second{0.990} \\
Mondrian conformal & \best{0.965} & 0.947 & \best{0.933} & 0.956 & 0.957 & 0.928 & \second{0.990} \\
Localized calibration & 0.676 & \second{0.965} & 0.867 & 0.986 & 0.988 & 0.965 & \best{1.000} \\
Localized calibration, mean support & 0.678 & 0.964 & \second{0.869} & \second{0.986} & 0.989 & 0.964 & \best{1.000} \\
Localized calibration, no semantic term & 0.676 & 0.963 & 0.849 & 0.986 & 0.989 & 0.968 & \best{1.000} \\
\hline
\end{tabular}
\caption{Class-wise accepted coverage on the ISIC 2019 external-shift evaluation using frozen BiomedCLIP. Class counts are AKIEC: 374, BCC: 975, BKL: 660, DF: 91, NV: 2495, MEL: 1327, and VASC: 104.}
\label{tab:supp-isic2019-classwise}
\end{table}

\begin{table}[H]
\centering
\footnotesize
\setlength{\tabcolsep}{3.5pt}
\begin{tabular}{lccccccc}
\hline
Method & AKIEC & BCC & BKL & DF & NV & MEL & VASC \\
\hline
Split conformal & 0.600 & 0.941 & 0.906 & 1.000 & 0.987 & 0.974 & 0.875 \\
Local conformal & 0.615 & 0.941 & 0.830 & 1.000 & 0.992 & 0.952 & 1.000 \\
Mondrian conformal & 1.000 & 0.950 & 0.969 & 1.000 & 0.955 & 0.930 & 1.000 \\
Localized calibration & 0.688 & 0.967 & 0.859 & 1.000 & 0.987 & 0.959 & 1.000 \\
Localized calibration, mean support & 0.688 & 0.967 & 0.873 & 1.000 & 0.989 & 0.955 & 1.000 \\
Localized calibration, no semantic term & 0.688 & 0.967 & 0.840 & 1.000 & 0.991 & 0.959 & 1.000 \\
\hline
\end{tabular}
\caption{Class-wise accepted coverage on the HAM10000 in-distribution evaluation using frozen BiomedCLIP. Class counts are AKIEC: 65, BCC: 101, BKL: 223, DF: 20, NV: 1344, MEL: 227, and VASC: 24.}
\label{tab:supp-ham10000-classwise}
\end{table}

%% file: references.bib
@inproceedings{radford2021learning,
  title = {Learning Transferable Visual Models From Natural Language Supervision},
  author = {Radford, Alec and others},
  booktitle = {Proceedings of the 38th International Conference on Machine Learning},
  series = {Proceedings of Machine Learning Research},
  volume = {139},
  pages = {8748--8763},
  publisher = {PMLR},
  year = {2021},
  url = {https://proceedings.mlr.press/v139/radford21a.html}
}

@inproceedings{wang2022medclip,
  title = {{MedCLIP}: Contrastive Learning from Unpaired Medical Images and Text},
  author = {Wang, Zifeng and Wu, Zhenbang and Agarwal, Dinesh and Sun, Jimeng},
  booktitle = {Proceedings of the 2022 Conference on Empirical Methods in Natural Language Processing},
  pages = {3876--3887},
  year = {2022},
  doi = {10.18653/v1/2022.emnlp-main.256}
}

@article{zhou2022coop,
  title = {Learning to Prompt for Vision-Language Models},
  author = {Zhou, Kaiyang and Yang, Jingkang and Loy, Chen Change and Liu, Ziwei},
  journal = {International Journal of Computer Vision},
  volume = {130},
  number = {9},
  pages = {2337--2348},
  year = {2022},
  doi = {10.1007/s11263-022-01653-1}
}

@article{zhang2023biomedclip,
  title = {{BiomedCLIP}: A Multimodal Biomedical Foundation Model Pretrained from Fifteen Million Scientific Image-Text Pairs},
  author = {Zhang, Sheng and others},
  journal = {arXiv preprint arXiv:2303.00915},
  year = {2023},
  doi = {10.48550/arXiv.2303.00915}
}

@inproceedings{xia2024cares,
  title = {{CARES}: A Comprehensive Benchmark of Trustworthiness in Medical Vision Language Models},
  author = {Xia, Peng and others},
  booktitle = {Advances in Neural Information Processing Systems},
  volume = {37},
  pages = {140334--140365},
  year = {2024},
  doi = {10.52202/079017-4455}
}

@inproceedings{xie2025medtrinity,
  title = {{MedTrinity-25M}: A Large-Scale Multimodal Dataset with Multigranular Annotations for Medicine},
  author = {Xie, Yunfei and others},
  booktitle = {International Conference on Learning Representations},
  year = {2025},
  url = {https://openreview.net/forum?id=IwgmgidYPS}
}

@inproceedings{liu2025kpl,
  title = {{KPL}: Training-Free Medical Knowledge Mining of Vision-Language Models},
  author = {Liu, Jiaxiang and others},
  booktitle = {Proceedings of the AAAI Conference on Artificial Intelligence},
  volume = {39},
  pages = {18852--18860},
  year = {2025},
  doi = {10.1609/aaai.v39i18.34075}
}

@inproceedings{koleilat2025biomedcoop,
  title = {{BiomedCoOp}: Learning to Prompt for Biomedical Vision-Language Models},
  author = {Koleilat, Taha and Asgariandehkordi, Hojat and Rivaz, Hassan and Xiao, Yiming},
  booktitle = {Proceedings of the IEEE/CVF Conference on Computer Vision and Pattern Recognition},
  pages = {14766--14776},
  year = {2025}
}

@inproceedings{gutbrod2025openmibood,
  title = {{OpenMIBOOD}: Open Medical Imaging Benchmarks for Out-of-Distribution Detection},
  author = {Gutbrod, Max and Rauber, David and Nunes, Danilo Weber and Palm, Christoph},
  booktitle = {Proceedings of the IEEE/CVF Conference on Computer Vision and Pattern Recognition},
  pages = {25874--25886},
  year = {2025}
}

@inproceedings{basu2025calibprompt,
  title = {Calibration-Aware Prompt Learning for Medical Vision-Language Models},
  author = {Basu, Abhishek and Shamshad, Fahad and Sharifdeen, Ashshak and Nandakumar, Karthik and Khan, Muhammad Haris},
  booktitle = {36th British Machine Vision Conference},
  publisher = {BMVA},
  year = {2025},
  url = {https://bmva-archive.org.uk/bmvc/2025/assets/papers/Paper_1062/paper.pdf}
}

@inproceedings{silvarodriguez2025confot,
  title = {Conformal Prediction for Zero-Shot Models},
  author = {Silva-Rodriguez, Julio and Ben Ayed, Ismail and Dolz, Jose},
  booktitle = {Proceedings of the IEEE/CVF Conference on Computer Vision and Pattern Recognition},
  pages = {19931--19941},
  year = {2025}
}

@inproceedings{eslami2023pubmedclip,
  title = {{PubMedCLIP}: How Much Does {CLIP} Benefit Visual Question Answering in the Medical Domain?},
  author = {Eslami, Sedigheh and Meinel, Christoph and de Melo, Gerard},
  booktitle = {Findings of the Association for Computational Linguistics: EACL 2023},
  pages = {1181--1193},
  publisher = {Association for Computational Linguistics},
  year = {2023},
  doi = {10.18653/v1/2023.findings-eacl.88}
}

@misc{google2025medsiglip,
  title = {{MedSigLIP}: Medical Image-Text Encoder for Zero-Shot Classification and Retrieval},
  author = {{Google Health AI Developer Foundations}},
  year = {2025},
  howpublished = {\url{https://developers.google.com/health-ai-developer-foundations/medsiglip}},
  note = {Model card}
}

@article{cheng2025robustness,
  title = {Understanding the Robustness of Vision-Language Models to Medical Image Artefacts},
  author = {Cheng, Zijie and others},
  journal = {npj Digital Medicine},
  volume = {8},
  pages = {727},
  year = {2025},
  doi = {10.1038/s41746-025-02108-w}
}

@inproceedings{bozorgtabar2026lata,
  title = {{LATA}: Laplacian-Assisted Transductive Adaptation for Conformal Uncertainty in Medical {VLMs}},
  author = {Bozorgtabar, Behzad and others},
  booktitle = {Proceedings of the IEEE/CVF Conference on Computer Vision and Pattern Recognition},
  pages = {36311--36320},
  year = {2026}
}

@inproceedings{fuchs2026lcpvlm,
  title = {Localized Conformal Prediction for Image Classification with Vision-Language Models},
  author = {Fuchs, Cl{\'e}ment and Bary, Tim and Macq, Beno{\^i}t},
  booktitle = {2025 13th European Workshop on Visual Information Processing},
  pages = {1--6},
  publisher = {IEEE},
  year = {2025},
  doi = {10.1109/EUVIP66349.2025.11238757}
}

@inproceedings{liu2026tacp,
  title = {Conformal Prediction Meets Long-Tail Classification},
  author = {Liu, Shuqi and Huang, Jianguo and Ong, Luke},
  booktitle = {Proceedings of the AAAI Conference on Artificial Intelligence},
  volume = {40},
  pages = {23828--23836},
  year = {2026},
  doi = {10.1609/aaai.v40i28.39558}
}

@article{zeng2026cpr,
  title = {Empirical Bayes Conformal Prediction for Vision and Language Models},
  author = {Zeng, Jiapeng and Prabhu, Yogesh and Zeng, Zhanpeng and Newton, Michael A. and Singh, Vikas},
  journal = {arXiv preprint arXiv:2605.23189},
  year = {2026},
  doi = {10.48550/arXiv.2605.23189}
}

@inproceedings{miao2026biodpp,
  title = {{BioDPP}: Dynamic Prompt Policy Learning for Biomedical Vision-Language Models},
  author = {Miao, Pingyi and others},
  booktitle = {Proceedings of the AAAI Conference on Artificial Intelligence},
  volume = {40},
  pages = {8052--8060},
  year = {2026},
  doi = {10.1609/aaai.v40i10.37751}
}

@inproceedings{shao2026vmfcoop,
  title = {{vMFCoOp}: Towards Equilibrium on a Unified Hyperspherical Manifold for Prompting Biomedical {VLMs}},
  author = {Shao, Minye and others},
  booktitle = {Proceedings of the AAAI Conference on Artificial Intelligence},
  volume = {40},
  pages = {8851--8859},
  year = {2026},
  doi = {10.1609/aaai.v40i11.37839}
}

@inproceedings{cui2026biomedccpl,
  title = {{BiomedCCPL}: Causal Conditional Prompt Learning for Biomedical Vision-Language Models},
  author = {Cui, Xueliang and others},
  booktitle = {Proceedings of the IEEE/CVF Conference on Computer Vision and Pattern Recognition},
  pages = {40812--40821},
  year = {2026}
}

@inproceedings{mahapatra2026valiant,
  title = {{VALIANT}: Prompt Instability for Active Learning in Black-Box Medical Imaging},
  author = {Mahapatra, Dwarikanath and Bozorgtabar, Behzad and Roy, Sudipta and Razzak, Imran and Reyes, Mauricio},
  booktitle = {Proceedings of the AAAI Conference on Artificial Intelligence},
  volume = {40},
  pages = {7901--7909},
  year = {2026},
  doi = {10.1609/aaai.v40i10.37734}
}

@inproceedings{guo2017calibration,
  title = {On Calibration of Modern Neural Networks},
  author = {Guo, Chuan and Pleiss, Geoff and Sun, Yu and Weinberger, Kilian Q.},
  booktitle = {Proceedings of the 34th International Conference on Machine Learning},
  series = {Proceedings of Machine Learning Research},
  volume = {70},
  pages = {1321--1330},
  publisher = {PMLR},
  year = {2017},
  url = {https://proceedings.mlr.press/v70/guo17a.html}
}

@book{vovk2005algorithmic,
  title = {Algorithmic Learning in a Random World},
  author = {Vovk, Vladimir and Gammerman, Alexander and Shafer, Glenn},
  publisher = {Springer},
  year = {2005}
}

@inproceedings{angelopoulos2021uncertainty,
  title = {Uncertainty Sets for Image Classifiers Using Conformal Prediction},
  author = {Angelopoulos, Anastasios N. and Bates, Stephen and Jordan, Michael I. and Malik, Jitendra},
  booktitle = {International Conference on Learning Representations},
  year = {2021}
}

@inproceedings{ovadia2019can,
  title = {Can You Trust Your Model's Uncertainty? Evaluating Predictive Uncertainty Under Dataset Shift},
  author = {Ovadia, Yaniv and others},
  booktitle = {Advances in Neural Information Processing Systems},
  volume = {32},
  year = {2019}
}

@inproceedings{tibshirani2019conformal,
  title = {Conformal Prediction Under Covariate Shift},
  author = {Tibshirani, Ryan J. and Barber, Rina Foygel and Candes, Emmanuel J. and Ramdas, Aaditya},
  booktitle = {Advances in Neural Information Processing Systems},
  volume = {32},
  pages = {2526--2536},
  year = {2019}
}

@article{guan2022bcops,
  title = {Prediction and Outlier Detection in Classification Problems},
  author = {Guan, Leying and Tibshirani, Robert},
  journal = {Journal of the Royal Statistical Society Series B: Statistical Methodology},
  volume = {84},
  number = {2},
  pages = {524--546},
  year = {2022},
  doi = {10.1111/rssb.12443}
}

@article{guan2023localized,
  title = {Localized Conformal Prediction: A Generalized Inference Framework for Conformal Prediction},
  author = {Guan, Leying},
  journal = {Biometrika},
  volume = {110},
  number = {1},
  pages = {33--50},
  year = {2023},
  doi = {10.1093/biomet/asac040}
}

@article{hore2025localweights,
  title = {Conformal Prediction with Local Weights: Randomization Enables Robust Guarantees},
  author = {Hore, Rohan and Barber, Rina Foygel},
  journal = {Journal of the Royal Statistical Society Series B: Statistical Methodology},
  volume = {87},
  number = {2},
  pages = {549--578},
  year = {2025},
  doi = {10.1093/jrsssb/qkae103}
}
